\documentclass{article}
\usepackage{fullpage}
\usepackage{lineno,hyperref}
\modulolinenumbers[5]

\usepackage[utf8]{inputenc}
\usepackage{caption}
\usepackage{soul}
\usepackage{mathtools}
\usepackage{caption}
\usepackage[ruled,linesnumbered]{algorithm2e}

\usepackage{multirow}
\usepackage{epsfig}
\usepackage{epstopdf}
\usepackage{amsmath}
\usepackage{amssymb}
\usepackage{subfig}
\usepackage{soul,color}

\usepackage[noend]{algorithmic}
\usepackage{url}
\makeatletter
\DeclareRobustCommand*\cal{\@fontswitch\relax\mathcal}
\makeatother
\newcommand{\squishlist}{
 \begin{list}{$\bullet$}
  { \setlength{\itemsep}{0pt}
     \setlength{\parsep}{2pt}
     \setlength{\topsep}{2pt}
     \setlength{\partopsep}{0pt}
     \setlength{\leftmargin}{1em}
     \setlength{\labelwidth}{1em}
     \setlength{\labelsep}{0.5em} } }

     \newcommand{\squishend}{
  \end{list}  }

\usepackage{graphicx}

\newtheorem{prop}{Proposition}

\newtheorem{lemm}{Lemma}
\newtheorem{corollary-1}{Corollary}
\newtheorem{Theorem}{Theorem}
\newtheorem{Example}{Example}

\def\multiset#1#2{\ensuremath{\left(\kern-.3em\left(\genfrac{}{}{0pt}{}{#1}{#2}\right)\kern-.3em\right)}}

\title{Competitive Ratios for Online Multi-capacity Ridesharing}

\author{Meghna Lowalekar, Pradeep Varakantham, Patrick Jaillet$^\dagger$\\}
\date{\small{School of Information Systems, Singapore Management University\\
       $\dagger$ Department of Electrical Engineering and Computer Science, Massachussets Institute of Technology\\
       meghnal.2015@phdcs.smu.edu.sg, pradeepv@smu.edu.sg, jaillet@mit.edu}}

\begin{document}
\maketitle
\begin{abstract}
In multi-capacity ridesharing, multiple requests (e.g., customers, food items, parcels) with different origin and destination pairs travel in one resource. In recent years, online multi-capacity ridesharing services (i.e., where assignments are made online) like Uber-pool, foodpanda, and on-demand shuttles have become hugely popular in transportation, food delivery, logistics and other domains. This is because multi-capacity ridesharing services benefit all parties involved -- the customers (due to lower costs), the drivers (due to higher revenues) and the matching platforms (due to higher revenues per vehicle/resource). Most importantly these services can also help reduce carbon emissions (due to fewer vehicles on roads).

Online multi-capacity ridesharing is extremely challenging as the underlying matching graph is no longer bipartite (as in the unit-capacity case) but a tripartite graph with resources (e.g., taxis, cars), requests and request groups (combinations of requests that can travel together). The desired matching between resources and request groups is constrained by the edges between requests and request groups in this tripartite graph (i.e., a request can be part of at most one request group in the final assignment). While there have been myopic heuristic approaches employed for solving the online multi-capacity ridesharing problem, they do not provide any guarantees on the solution quality.  
 

To that end, this paper presents the first approach with bounds on the competitive ratio for online multi-capacity ridesharing (when resources rejoin the system at their initial location/depot after serving a group of requests). The competitive ratio is : (i) 0.31767 for capacity 2; and (ii) $\gamma$ for any general capacity $\kappa$, where $\gamma$ is a solution to the equation $\gamma = (1-\gamma)^{\kappa+1}$.
\end{abstract}


\section{Introduction}
\noindent Motivated by multiple online to offline services including point-to-point transportation, food delivery, logistics, etc., online matching problems have received tremendous interest in the recent years. Specifically, on-demand unit-capacity (e.g., UberX, Lyft) and multi-capacity (e.g., Uberpool, Lyftline, Deliveroo, Food Panda) ridesharing services have become hugely popular in many cities around the world. In these platforms, resources have to be matched online (in real-time) to either one request (unit-capacity) or a group of requests (multi-capacity) so as to maximize the weight of the matching (e.g., revenue, number of requests served). 

Given the win-win properties of multi-capacity ridesharing to all the concerned parties (customers, drivers, matching platform) and the environment (through reduced carbon emissions), we are interested in developing a {performance guaranteed approach for multi-capacity ridesharing}. 

There are two major threads of relevant research. The {\em first thread} is on online unit-capacity ridesharing where the underlying problem is an online bipartite matching problem. The standard online bipartite matching problem involves matching known (i.e., available offline) disposable resources~\footnote{Once a resource is assigned, it can not be used by any other incoming vertex/request.} on one side to the online arriving vertices/requests on the other side, over multiple timesteps. Many approaches provide performance guarantees under different arrival assumptions for incoming vertices~\cite{karp1990optimal,devanur2013randomized,jaillet2013online}. Mehta~\cite{mehta2013online} provides a detailed survey of the same. One popular arrival assumption is the known identical independent distribution (KIID)~\cite{jaillet2013online,manshadi2012online}, where online vertices arrive over $T$ rounds and their arrival distributions are assumed to be identically distributed and independent over $T$ rounds. This distribution is also known to the online algorithm in advance. The existing literature provide bounds of at least $1-\frac{1}{e}$ on the expected competitive ratio (ratio of the expected value obtained by the algorithm to the expected value obtained by an offline optimal algorithm) for online bipartite matching problems under KIID. 

In case of unit-capacity ridesharing, the offline available resources (i.e., vehicles) are reusable. Dickerson {\em et.al.}~\cite{dickerson2017allocation} were able to provide a $\frac{1}{2}$ bound for the unit-capacity ridesharing in which resources are reusable and they join the system after serving the requests at the same location. Instead of KIID, they consider that arrival distributions of online vertices can change from time to time (i.e., it is not iid) but this distribution is also known to the algorithm. They refer to this distribution as the Known Adversarial Distribution (KAD).  

Unfortunately, this thread of work is only applicable for unit-capacity resources and cannot be directly adapted to consider multi-capacity resources because the underlying problem is no longer an online bipartite matching problem (see below). Another limitation is that the existing work for unit-capacity ridesharing has primarily focused on requests arriving sequentially (i.e., one by one) and not in batches which is a desirable property when considering multi-capacity ridesharing problems (for instance, last mile services at train stations need to consider that the large number of passengers will arrive and request for last mile transportation to their home at the same time.).

The {\em second thread} of relevant research is on approaches to solve online multi-capacity (capacity $> $ 1) ridesharing problems. There have been multiple heuristic approaches~\cite{alonso2017demand,lowalekarVJ19} provided for solving the ridesharing problem for multi-capacity resources in batch arrival model. However, none of these approaches provide any bounds on the performance and are typically myopic (i.e., they do not consider any future information) due to the challenging nature of the problem. 

The multi-capacity resources (capacity $>$ 1) make the problem challenging because resources have to be matched to groups of requests and not just to individual requests. This results in a significant change in the structure of the underlying matching graph. Unlike unit-capacity ridesharing, where the underlying graph is bipartite, the multi-capacity ridesharing has a tripartite graph~\cite{beineke1980four} with reusable resources (vehicles), request groups (i.e., combinations of passenger requests) and online vertices (corresponding to passenger requests). The desired matching between the resources and request groups (combination of requests) is constrained by the edges between requests and request groups (i.e., a request can be part of at most one request group in final assignment) in this tripartite graph. It should be noted that this matching problem in tripartite graph is not equivalent to any variant of bipartite matching problem~\cite{aggarwal2011online,feldman2009online,lee2017maximum,huang2018match} studied in the literature. This is because the weight of a match and the time after which resource becomes available again is dependent on the requests which are paired together in the group assigned to the resource.  
 
To the best of our knowledge, there has been no research on providing performance guaranteed algorithms for such tripartite graphs. There has been some work on solving a part of this matching problem which focused on finding the requests which can be grouped together over time by considering the sequential arrival of requests~\cite{ashlagi2018maximum,ashlagi2019edge} in the adversarial and random order arrival. However, these works ignore the main component of matching the resources to the request groups.

\subsection{Contributions}
{Our \textit{first} contribution} is in designing a performance guaranteed online algorithm that provides a competitive ratio of $\frac{1}{2}$ for the unit-capacity ridesharing problem that considers batch arrival of online vertices~\footnote{The online arriving vertices correspond to the requests. Throughout the paper we use vertices and requests interchangeably.} under the known arrival distribution. Due to the change in the value obtained by optimal algorithm (more details in Section \ref{sect:extension}), it is not obvious whether the competitive ratio will increase or decrease or remain the same as compared to the sequential arrival case~\cite{dickerson2017allocation}. Therefore, this is an important result where in we are able to show that the same competitive ratio can be achieved even when the vertices arrive in batches.

\noindent {Our \textit{second and the main} contribution} is to provide a performance guaranteed online algorithm that provides a non-zero competitive ratio for the online multi-capacity ridesharing problems considering batch arrival of online vertices under the known arrival distribution. The competitive ratio is:
\squishlist
\item 0.31767 for capacity 2 
\item $\gamma$ for any arbitrary capacity $\kappa$, where $\gamma$ is solution to the the expression $(1-\gamma)^{\kappa+1} = \gamma$. 
\squishend
{ \em Even though we require groups of vertices in this online algorithm, these groups can be generated offline and hence does not add to the run-time complexity. } These general bounds for arbitrary capacity ridesharing are applicable under the assumption that the type of the resources/vehicles (i.e., their location) rejoining the system (after serving a group of vertices) does not change~\cite{dickerson2017allocation}. 

\noindent {\em Finally}, we provide simple heuristics (based on the offline optimal LP) which work well in practice (as demonstrated in our experimental results).


\section{Background}
\label{sect:bg}

In this section, we provide the formal definition of expected competitive ratio and the research relevant~\cite{dickerson2017allocation} to the work in this paper. 

\subsection{Expected Competitive Ratio}
The performance of any online algorithm is measured using a metric called competitive ratio. An online algorithm with a competitive ratio of $\gamma$ is called $\gamma$-competitive algorithm. In case of known distribution models, the expected value of the competitive ratio is employed. The expected competitive ratio of any algorithm ALG is defined~\cite{mehta2013online} as $\min_{I,D} \frac{E[ALG(I,D)]}{E[OPT(I)]}$, where $I$ denotes the input and $D$ denotes the arrival distribution and $E[OPT(I)]$ denotes the expected value of the offline optimal algorithm. In general, an upper bound on the value of $E[OPT(I)]$ is provided by using a benchmark linear program. This results in providing a valid lower bound on the resulting competitive ratio.  

\noindent {\em Since we only employ expected competitive ratio in this paper, we henceforth just refer to it as competitive ratio.}

\subsection{OM-RR-KAD}
We now describe the Online Matching with (Offline) Reusable Resources under Known Adversarial Distributions (OM-RR-KAD) model ~\cite{dickerson2017allocation} for ridesharing in which the vehicle capacity is restricted to 1. OM-RR-KAD is a bipartite matching problem between offline reusable resources (e.g., vehicles), ${\cal U}$, and vertices that arrive online, ${\cal V}$ (e.g., user requests), over $T$ rounds~\footnote{We use round and timestep interchangeably in the paper}. Online vertices arrive according to a {\em Known Adversarial Distribution (KAD)} represented by a set of arrival probabilities, $\{p_{v}^t\}$ ($\sum_{v} p_{v}^t = 1, \forall t$). Once an online vertex of type $v$ arrives (i.e., sampled from $p_v^t$), an irrevocable decision needs to be taken immediately to match it to one of the offline resources, for which a weight, $w_{u,v}^{t}$ is received, or to reject it. The offline resource becomes unavailable for a few rounds after it is matched and the number of rounds of unavailability, $c_{u,v}^{t}$, is characterized by an integral distribution, $c_{u,v}^t \in \{1,2,\ldots,T\}$. The offline resource rejoins the system after $c_{u,v}^{t}$ rounds. The goal is to design an online assignment policy that will maximize the weight. 


There are two key steps in obtaining a performance guaranteed online assignment policy:
\begin{table}[t]
\center
        \begin{tabular}{|r|}
        \hline

        \begin{minipage}{0.95\textwidth}
                \vspace{0.05in}
\textbf{LPSequential:}
{
\begingroup
\addtolength{\jot}{-2pt}
{ \begin{align}
\max & \sum_{t =0}^{T} \sum_{u \in {\cal U}} \sum_{v \in {\cal V}} w_{u,v}^{t} \cdot x_{u,v}^{t} \nonumber \\
s.t. \quad & \sum_{u \in {\cal U}} x_{u,v}^{t} \leq p_{v}^{t} ::: \forall v \in {\cal V}, 0 \leq t < T \label{cons:opt21} \\
& \sum_{t'=0}^{t} \sum_{v' \in {\cal V}} x_{u,v'}^{t'} \cdot Pr[c_{u,v'}^{t'} > t-t'] + \sum_{v \in {\cal V}} x_{u,v}^{t} \leq 1 \nonumber  \\ 
&\hspace{1in} ::: \forall u \in {\cal U}, 0 \leq t < T \label{cons:opt22}\\
& 0 \leq x_{u,v}^{t} \leq 1 ::: \forall u \in {\cal U}, v \in {\cal V}, 0 \leq t < T 
\end{align}}
\vspace{-8pt}
\endgroup }
\end{minipage} \\
        \hline
        \end{tabular}
        \caption{{Unit Capacity Sequential Arrival}}
        \label{opt:seq}
        \end{table}


\noindent \textbf{\em First}, an upper bound on the offline optimal, $\textbf{x}^*$ is computed using the linear program (LP) of Table~\ref{opt:seq}. $x^{*,t}_{u,v}$ denotes the probability of assigning resource $u$ to online vertex of type $v$ in round $t$. Constraint \eqref{cons:opt21} ensures that the expected number of times a vertex of type $v$ is matched is less than or equal to the expected number of times the vertex is available. Constraint \eqref{cons:opt22} ensures that the resource $u$ is assigned in round $t$ if and only if it is available in round $t$. It should be noted that this LP provides a solution over all realizations of online vertices and hence that solution may not be applicable to a specific instantiation of online vertex (as the corresponding $u$ may not be available). 

\noindent \textbf{\em Second}, an assignment rule is provided to compute the online probability of assigning a resource $u$ for a specific instantiation of online vertex (of type $v$ in round $t$) and is given by:
{
\begin{align}
\frac{x^{*,t}_{u,v} \cdot \gamma}{p_{v}^t \cdot \beta_{u}^t} \label{eqn1}
\end{align}
}
\noindent where 
$x^{*,t}_{u,v}$ is a solution to the LP in Table \ref{opt:seq} and $\gamma$ is the desired competitive ratio of the online assignment; and 
$\beta_{u}^t$ is the probability that resource $u$ is safe for assignment in round $t$. By simulating the current strategy up to $t$, $\beta_{u}^t$ can be estimated with a small error. \\

The following theorem characterizes the $\frac{1}{2}$ bound on the expected competitive ratio. 
\begin{Theorem}
Dickerson~{\em et.al.}[\cite{dickerson2017allocation}] The optimal value of LPSequential in Table~\ref{opt:seq} provides a valid upper bound on the offline optimal value for OM-RR-KAD. The online assignment rule of Equation~\ref{eqn1} based on the LP achieves an online competitive ratio of $\frac{1}{2}-\epsilon$ for any given $\epsilon > 0$.
\end{Theorem}

The $\epsilon$ factor comes in the competitive ratio due to the error in the estimation of $\beta_{u}^{t}$. For a clean presentation, throughout the paper, we assume that these values can be estimated correctly and ignore the estimation error. 

\section{Batch Arrival of vertices}
\label{sect:extension}
In ridesharing problems, user requests typically arrive in batches instead of arriving sequentially (e.g., users coming out of a train, theatre or mall looking for shared rides). So, we extend the OM-RR-KAD model to consider batch arrival of online vertices and also provide an online algorithm that achieves the same competitive ratio of $\frac{1}{2}$ as in the sequential arrival case. Batch arrival is different from sequential arrival because multiple online vertices (more information at each step) have to be matched to multiple offline resources at each round. 

Since there are more vertices available in each round, online algorithms can potentially make better assignments in the batch case as compared to the sequential case. Due to this, it seems that the competitive ratio in the batch arrival case will be higher than the sequential arrival case. However, 
\squishlist
\item As the assignment for any vertex should be made in the same round of its arrival, in batch case where each round has multiple vertices, optimal algorithm (denominator of competitive ratio) also considers a greater number of vertices in each round and hence optimal value can also improve (as compared to the optimal value for sequential case). 
\item Compared to the sequential case, more time is spent deliberating (since we must wait until end of batch to make assignments) and during that time no assignment will happen and hence the number of vertices assigned by the optimal algorithm can be lower. 
\squishend
Therefore, the relationship between the competitive ratio for the sequential and batch cases is non trivial. We now provide an algorithm which ensures that the competitive ratio in batch arrival case is equal to the sequential arrival case. 

We first mention the changes required in OM-RR-KAD model for the batch arrival case and then provide the performance guaranteed online algorithm for the unit-capacity case. \\

\noindent \textbf{Changes to OM-RR-KAD for Batch Case:} In the OM-RR-KAD model, at each round $t$, a single vertex is sampled using the probability $\{p_{v}^t\}$. However, in the batch extension, $b^t$ vertices arrive at each round and each of these $b^t$ vertices is sampled using the same probabilities $\{p_{v}^t\}$. The expected number of vertices of type $v$ arriving in round $t$ is $q_v^t$ and is given by: { $$q_v^t = b^t \cdot p_v^t $$ }\\


\begin{figure*}
 \centering
   \includegraphics[width=0.3\textwidth,height=3.0in]{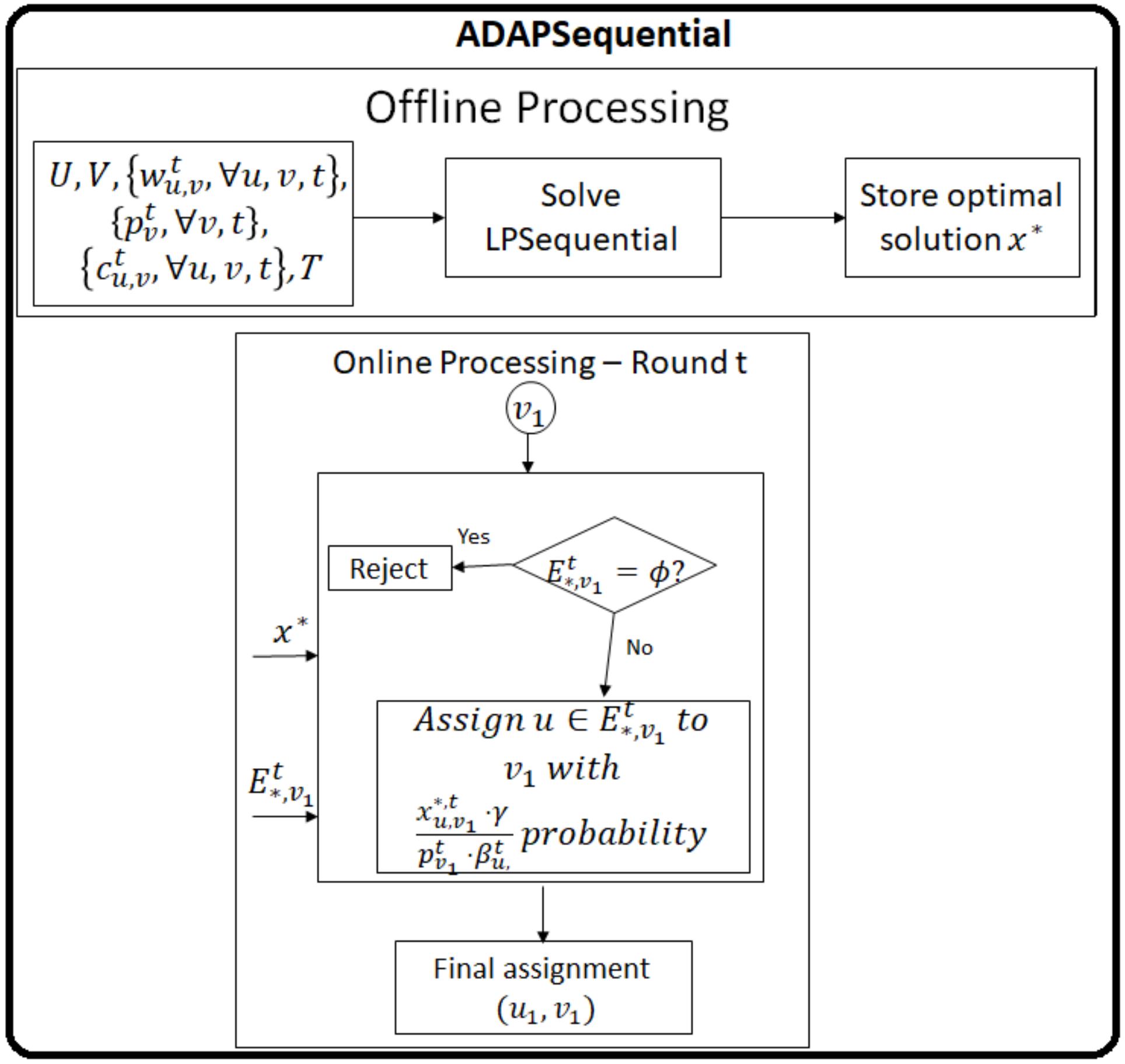}
   \includegraphics[width=0.55\textwidth,height=3.0in]{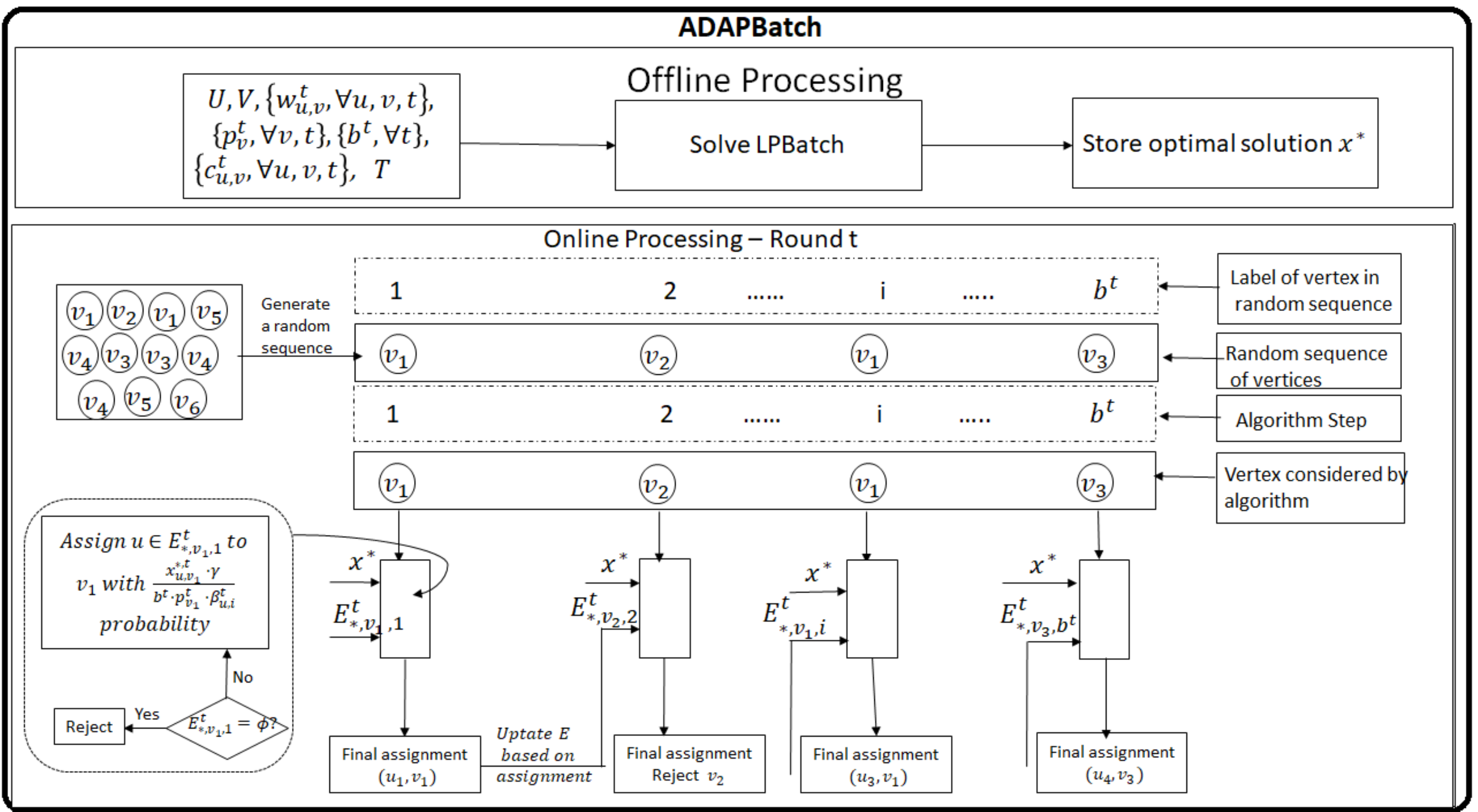}\\
   \includegraphics[width=0.7\textwidth,height=3.0in]{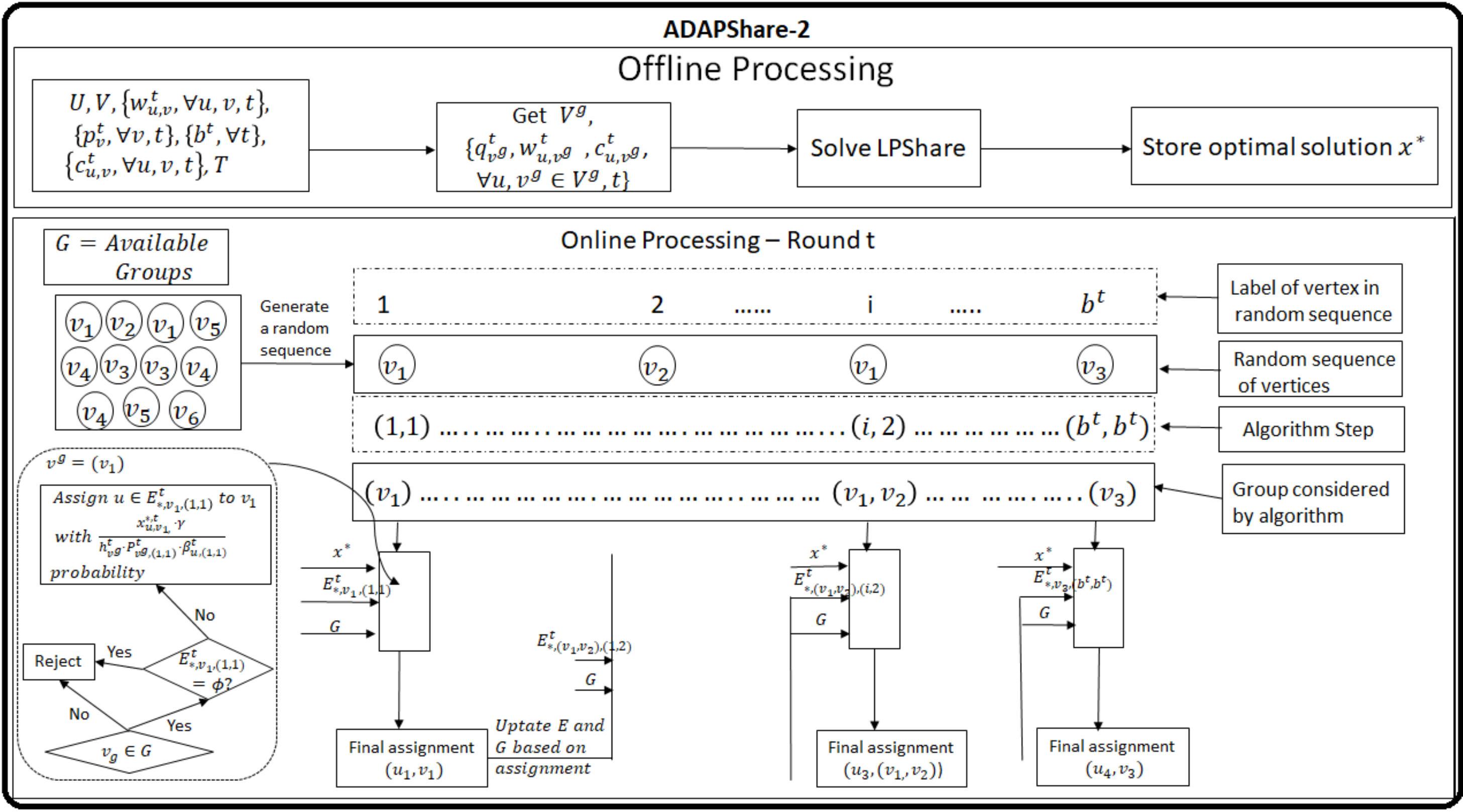}
\caption{{The Figure depicts the difference in the processing of algorithms in unit-capacity sequential, unit-capacity batch and multi-capacity share case. The online component in each of the algorithms corresponds to the processing in round $t$ for a single instance of arrival of vertices. We only show the detailed flow diagram in the first block for each of the algorithms, rest of the blocks will have similar flow.}}
  \label{fig:adap}
\end{figure*}

\noindent \textbf{LP for Upper Bound on Offline Batch Optimal, LPBatch:}
The optimization formulation for the batch case is same as the LP in Table~\ref{opt:seq}, except for the constraint in Equation~\eqref{cons:opt21}. Given that there are $q_{v}^t$ (and not $p_v^t$) expected arrivals of vertex of type $v$ at each round, the modified constraint is: 
{
\begin{align}
\sum_{u \in {\cal U}} x_{u,v}^{t} \leq q_{v}^{t} ::: \forall v \in {\cal V} , 0 \leq t < T \label{eqn:cons2}
\end{align}
}

\noindent We will refer to the modified LP as LPBatch. 

 \begin{prop}
\label{prop:lpbatch}
\noindent The optimal value of LPBatch provides a valid upper bound on the offline optimal value~\footnote{Proof is available in appendix.}.
\end{prop}
{
\begin{algorithm}
\caption{ADAPBatch($\gamma$)}
\begin{algorithmic}[1]
\FOR{$t < T$}
\STATE{Generate a random shuffling of the incoming $b^{t}$ vertices. Label the vertices from 1 to $b^{t}$.}
\FOR{$i=1$ to $b^{t}$}
\STATE{$v$ = type of vertex with label $i$}
\STATE{\textbf{If} ${E}_{*,v,i}^{t} = \phi$, then reject the vertex with label $i$;}
\STATE{\textbf{Else} choose $u \in {E}_{*,v,i}^{t}$ with probability $\frac{x_{u,v}^{*,t} \cdot \gamma}{q_{v}^{t} \cdot \beta_{u,i}^{t}}$}
\STATE{Update the sets ${E}_{*,v,j}^{t}$ for all $j > i$ based on the assignment.}
\ENDFOR
\ENDFOR
\end{algorithmic}
\label{alg:adapmod}
\end{algorithm}
}
\noindent \textbf{ADAPBatch}
The online algorithm presented in Algorithm \ref{alg:adapmod} is used to make an online assignment of the resources to the incoming vertices that are arriving in batches. We use an adaptive algorithm~\footnote{For an LP-based algorithm, we say that the algorithm is adaptive if for a given LP solution, the computation of strategy in each round $t$ depends on the strategies in the previous rounds~\cite{dickerson2019online}.} that employs the probability of a resource being safe (available for assignment) while making assignments.
The assignment rule to compute the online probability of assigning a resource $u$ for the vertex of type $v$ with label $i$ in round $t$ is:
{ $$\frac{x_{u,v}^{*,t} \cdot \gamma}{b^{t} \cdot p_{v}^{t} \cdot \beta_{u,i}^{t}}$$}
\noindent where $\beta_{u,i}^{t}$ denotes the probability that resource $u$ is safe in round $t$ when the vertex with label $i$ is being considered; and ${E}_{*,v,i}^{t} \subset {\cal U}$ is used to denote the set of safe neighbours for a vertex of type $v$ in round $t$ when the vertex with label $i$ is being considered.

In the algorithm, we process the vertices that have arrived in a batch one by one by considering a uniform random shuffling of incoming vertices. The intuition behind the assignment rule is to divide the optimal assignment for round $t$ uniformly into $b^{t}$ steps ($\frac{x_{u,v}^{*,t}}{b^{t}}$) and then to make sure that the vertex of type $v$ is matched to resource $u$ at any step with probability $\frac{x_{u,v}^{*,t} \cdot \gamma}{b^{t}}$ unconditionally. 
Another key change in the algorithm from the sequential case is the last step where the availability of offline resources is updated based on assignments made in the same round. 
Figure \ref{fig:adap} highlights the difference in the way the algorithms process online information in the sequential and batch case. 

\begin{prop}
\label{prop:adapbatch}
The online algorithm ADAPBatch is $\frac{1}{2}$ competitive.
\end{prop}
\noindent\textbf{Proof Sketch:} The maximum value of $\gamma$ for which the algorithm ADAPBatch is valid~\footnote{Algorithm is valid when the assignment rule probability lies between 0 and 1.} is $\gamma=\frac{1}{2}$. The proof involves showing that the minimum possible value of $\beta_{u,i}^{t}$ is $\frac{1}{2}$, for which we use mathematical induction. Finally, we show that ADAPBatch is $\gamma$ competitive and since the maximum value of $\gamma$ for which the assignment rule is valid is $\frac{1}{2}$, the algorithm is $\frac{1}{2}$ competitive. $\blacksquare$ 

\section{Multi-capacity Reusable Resources}

In this section, we provide a model, an online algorithm and competitive ratio analysis for the online multi-capacity ridesharing problem with reusable resources. 

\subsection{Model: OPERA}
\label{sect:model}
To address the challenges associated with multi-capacity resources, we propose a new model called OPERA ({\bf O}nline matching with offline multi-ca{\bf P}acity r{\bf E}usable {\bf R}esources in b{\bf A}tch Arrival Model).   
In OPERA, online vertices arrive in batches according to a {\em Known Adversarial Distribution (KAD)}. Once the online vertices arrive, there has to be an irrevocable decision made immediately on matching each offline resource $u$ to a group of online vertices $v^g$. The groups chosen for all vehicles should be such that each online vertex appears in at most one group. For each assignment of an offline resource $u$ to a group of online vertices of type $v^g$ in round $t$, a weight $w_{u, v^g}^t$ is received. After the assignment, the offline resource $u$ is unavailable for $c_{u,v^g}^t$ rounds before joining the system again~\footnote{In the context of last mile ridesharing -- after serving the group of passengers, vehicle comes back to its initial location}. The goal is to design an online assignment policy for assigning offline reusable resources to the groups of online vertices that will maximize the weight received over all time steps.

\begin{figure}
 \centering
   \includegraphics[width=0.6\textwidth,height=2.6in]{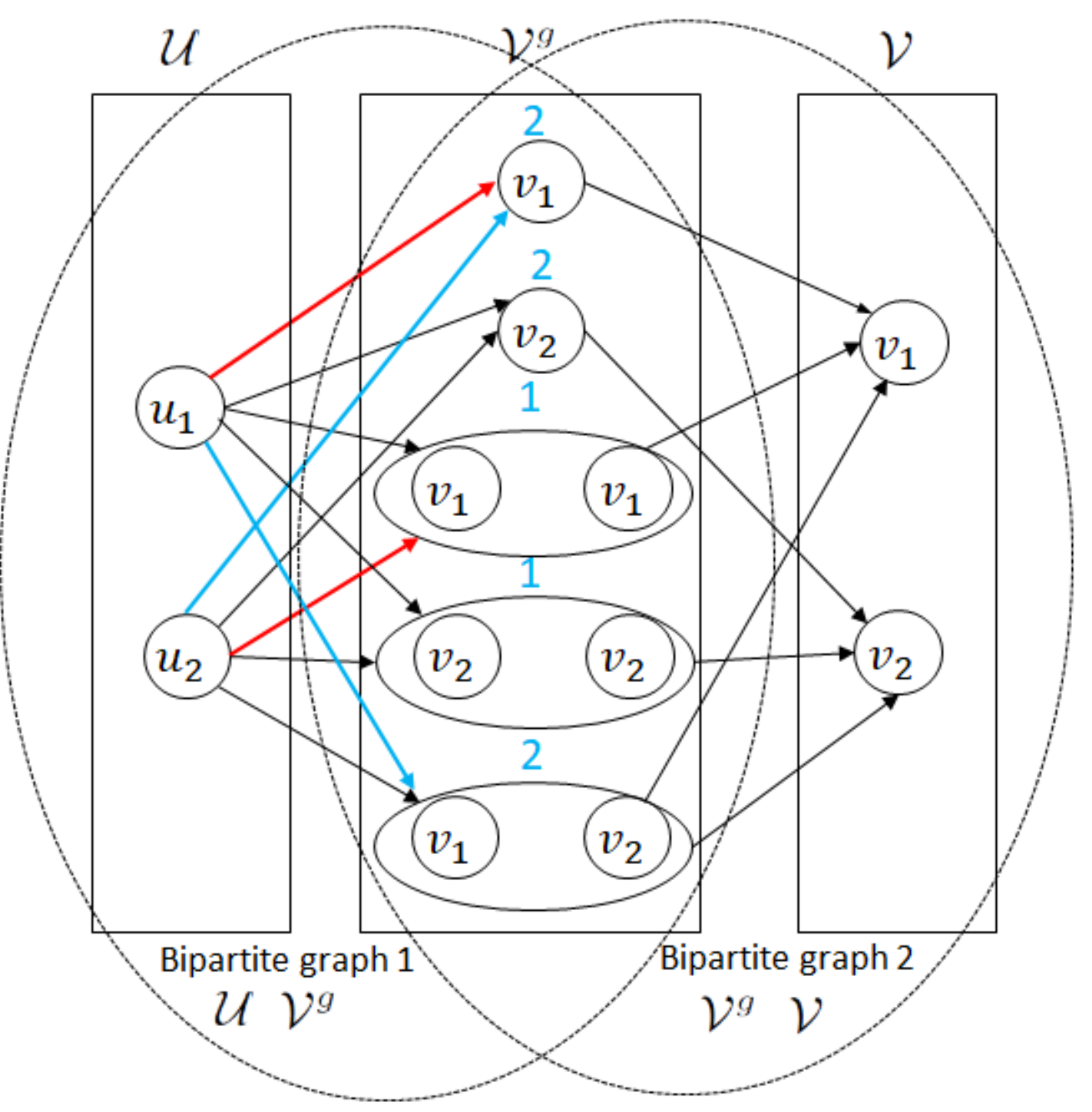}
\caption{{The Figure depicts the tripartite graph used in the OPERA model. It is a combination of 2 bipartite graphs. The goal is to find the matching in the first bipartite graph subject to the constraints enforced due to the edges present in the second bipartite graph. The blue numbers in ${\cal V}$ indicate the number of vertices of each type available and blue numbers in ${\cal V}^{g}$ denote the number of groups of each type which can be formed using available vertices in ${\cal V}$. The blue lines indicate a valid assignment of resources in ${\cal U}$ to groups in ${\cal V}^{g}$. Red lines indicate an invalid assignment as the vertex of type $v_{1}$ is used 3 times in this assignment but there are only 2 vertices of type $v_{1}$ available.}}
  \label{fig:operagraph}
\end{figure}
\noindent {\em Unlike in OM-RR-KAD, the underlying problem in OPERA is no longer a bipartite matching problem but a matching in a tripartite graph containing offline resources, ${\cal U}$ groups of online vertices, ${\cal V}^g$ and online vertices, ${\cal V}$.Figure~\ref{fig:operagraph} shows the tripartite graph formed in the case of OPERA.} \\

\noindent Here are other key differences between OPERA and OM-RR-KAD:
\squishlist
\item[${\cal U}$:] Each offline resource, $u \in {\cal U}$ in OPERA has a fixed capacity $\kappa$. 
\item[${\cal V}^g$:] As $\kappa > 1$, unlike in OM-RR-KAD model, resources can be assigned to more than one vertex at a round, i.e., resources can be assigned to groups of vertices where group sizes vary from 1 to $\kappa$. For ease of analysis, we consider that all the vertices can be paired together, and the constraints on the feasibility of pairing of vertices are handled through the weights received. Types of groups of vertices are obtained by generating all possible combinations (with repetitions) of size 1 to $\kappa$ of the set ${\cal V}$\footnote{\multiset{n}{k} denotes the number of multisets of cardinality $k$, with elements taken from a finite set of cardinality $n$.}. The resulting set is denoted by ${\cal V}^{g}$. Therefore,

{

$$|{\cal V}^{g}| = \sum\limits_{k=1}^{\kappa} \multiset{|{\cal V}|}{k} = \sum_{k=1}^{\kappa} \binom{|{\cal V}|+k-1}{k}$$ 
}
\noindent For each group of type $v^{g}$, $n_{v,v^{g}}$ denotes the number of times vertex of type $v \in {\cal V}$ is present in group of type $v^{g}$ (From the example Figure \ref{fig:operagraph}, for $v^{g} = (v_{1},v_{1})$, $n_{v_{1},v^{g}}$ will be 2 and for $v^{g} = (v_{1},v_{2})$, $n_{v_{1},v^{g}}$ will be 1.) 
\item[${q_v^t}$:] We consider batch arrival of vertices. Therefore, similar to the extension in Section~\ref{sect:extension}, $b^t$ vertices arrive at each round and each of these $b^t$ vertices is sampled using the same probabilities $\{p_{v}^t\}$. The expected number of vertices of type $v$ arriving in round $t$ is $q_v^t$ and is given by: {$$q_v^t = b^t \cdot p_v^t$$ }
\item[$w_{u,v^g}^t$:] Weight received is now based on the type of group assigned to the resource. 
\item[$c_{u,v^g}^t$:] Rounds of unavailability after an assignment is now based on the type of the group assigned to the resource. 
\squishend

Apart from the model differences, there are also differences with respect to the online assignments that can be made. The irrevocable assignment of resources in ${\cal U}$ to ${\cal V}^g$ should satisfy the following constraints: 
\squishlist
\item[\textbf{C1:}] Each resource $u \in {\cal U}$ is assigned at most once in each round.
\item[\textbf{C2:}] The total number of vertices of each type $v \in {\cal V}$ used in the assigned groups is less than or equal to the number of vertices available. 
\item[\textbf{C3:}] The number of groups of type $v^{g} \in {\cal V}^{g}$ assigned in round $t$ is less than or equal to the number of available groups of type $v^{g}$.

\squishend

\noindent In order to enforce constraint [C3] above in expectation (i.e., over all possible instantiations of arrivals), we need to compute $q_{v^{g}}^{t}$ --- the expected number of times group of type $v^g$ can be formed in round $t$. It is given by~\footnote{It corresponds to drawing $n_{v,v^{g}}$ vertices of each type $v \in v^{g}$ out of total $b^{t}$ trials for a multinomial distribution. Please refer to https://tinyurl.com/rjs524p for details on deriving the expression.}:
{ 
\begin{align}
q_{v^{g}}^{t} = h_{v^{g}}^{t} \prod\limits_{v \in v^{g}} (p_{v}^{t})^{n_{v,v^{g}}} \label{cons:hvg} \text{ where } h_{v^{g}}^{t} = \frac{\prod\limits_{i=0}^{i=|v^{g}|} (b^{t} -i)}{\prod\limits_{v \in {\cal V}}(n_{v,v^{g}})!}
\end{align}
}
We make the following assumptions in the model: (1) Once a resource $u$ is assigned to a group of type $v^{g}$ at $t$ it becomes unavailable for further matches for $c_{u,v^{g}}^{t}$ rounds irrespective of the size of $v^{g}$, i.e., insertion is not allowed. (2) The vertices can be grouped together iff they are arriving in same round. (3) For ease of explanation, we assume that $b^{t} > \kappa, \forall t$. However, this can be relaxed easily.

\subsection{Online Algorithm}

We first provide an LP for computing the upper bound on the offline optimal and then provide an adaptive assignment method based on the offline optimal solution. \\

\noindent \textbf{LP for Upper Bound on Offline Batch Optimal with Multi-Capacity Resources:} The optimization formulation~\footnote{LP is based on satisfying the flow constraints in the graph shown in Figure \ref{fig:operagraph}.}
is provided in Table \ref{opt:caphigh}. We refer to this LP as LPShare. Since LP is for the offline case over all possible instantiations on arrival vertices, the constraints hold in expectation. Constraints \eqref{cons:optc1}, \eqref{cons:optc2} and \eqref{cons:optc3} refer respectively to \textbf{C1}, \textbf{C2} and \textbf{C3} constraints (described in Section~\ref{sect:model}) in expectation (i.e., over all possible instantiations of arrivals). Constraint \eqref{cons:optc1} ensures that the resource $u$ is assigned in round $t$ iff $u$ is available in round $t$. 
 

\begin{table}[t]
\center
        \begin{tabular}{|r|}
        \hline

        \begin{minipage}{0.95\textwidth}
                \vspace{0.05in}
\textbf{LPShare:}
{
\begingroup
\addtolength{\jot}{-2pt}
\begin{align}
\max & \sum_{t =0}^{T} \sum_{u \in {\cal U}} \sum_{v^{g} \in {\cal V}^{g}} w_{u,v^{g}}^{t} \cdot x_{u,v^{g}}^{t} \\
s.t. \quad & \sum_{t'<t} \sum_{v^{g'} \in {{\cal V}^{g}}} x_{u,v^{g'}}^{t'} \cdot Pr[c_{u,v^{g'}}^{t'} > t-t'] + \nonumber\\
& \hspace{0.4in}+ \sum_{v^{g} \in {\cal V}^{g}} x_{u,v^{g}}^{t} \leq 1 ::: \forall u \in {\cal U}, 0 \leq t < T \label{cons:optc1}\\
& \sum_{v^{g}; v \in v^{g}} \sum_{u \in {\cal U}} n_{v,v^{g}} \cdot x_{u,v^{g}}^{t} \leq q_{v}^{t} ::: \forall v \in {\cal V}, 0 \leq t < T \label{cons:optc2}\\
& \sum_{u \in {\cal U}} x_{u,v^{g}}^{t} \leq q_{v^{g}}^{t} ::: \forall v^{g} \in {\cal V}^{g}, 0 \leq t < T \label{cons:optc3}\\
& 0 \leq x_{u,v^{g}}^{t} \leq 1 ::: \forall u \in {\cal U} ,v^{g} \in {\cal V}^{g}, 0 \leq t < T \label{cons:optc4}
\end{align}
\vspace{-8pt}
\endgroup }
\end{minipage} \\
        \hline
        \end{tabular}

        \caption{{Optimization Formulation - Multi-capacity}}
        \label{opt:caphigh}
        \end{table}
\begin{prop}
\label{prop:lpshare}
The optimal value of LPShare provides a valid upper bound on the offline optimal value~\footnote{Proof is available in appendix.}.
\end{prop}

{
\begin{algorithm}
\caption{ADAPShare-2($\gamma$)}
\begin{algorithmic}[1]
\FOR{$t < T$}
\STATE{Generate a random shuffling of the incoming $b^{t}$ vertices. Label the vertices from 1 to $b^{t}$.}
\FOR{$i=1$ to $b^{t}$}
\FOR{$j=1$ to $b^{t}$}
\STATE{$v^{g}$ = type of group formed at step $(i,j)$ based on the labels assigned to the vertices.}
\IF{$v^{g}$ is available for assignment at step $(i,j)$}
\STATE{\textbf{If} ${E}_{*,v^{g},(i,j)}^{t} == \phi$, reject $v^{g}$}
\STATE{\textbf{Else} choose $(u,v^{g}) \in {E}_{*,v^{g},(i,j)}^{t}$ with probability $p$ where $p= \frac{x_{u,v^{g}}^{*,t} \cdot \gamma}{h_{v^{g}}^{t} \cdot P_{v^{g},(i,j)}^{t} \cdot \beta_{u,(i,j)}^{t}}$}
\STATE{Update ${E}_{*,*,(i,j)}^{t}$, available groups based on the assignment.}
\ENDIF
\ENDFOR
\ENDFOR
\ENDFOR
\end{algorithmic}
\label{alg:suggestmatch}
\end{algorithm}
}

\noindent \textbf{ADAPShare-$\kappa$:}
For ease of explanation, we first present the online algorithm and competitive analysis for $\kappa = 2$. 

\noindent Let $x_{u,v^{g}}^{*,t}$ denotes the optimal probability of assigning a resource $u$ to a group of type $v^g$ in round $t$ (computed from offline optimal LP). We use Algorithm \ref{alg:suggestmatch} to make online assignment of resources to the groups of vertices based on $\{x_{u,v^{g}}^{*,t} \}$ values from the offline optimal LP. As shown in the algorithm, we perform a random shuffling of the $b^t$ vertices (that arrive in a batch in round $t$) and label the vertices from $1$ to $b^{t}$. The assignment of resources to groups is performed across $b^{t}\cdot b^{t}$ steps (as we consider groups of size 2). Step $(i,j)$ corresponds to a step where we compute the probability for assignment of a group formed by vertices with labels $i$ and $j$. It should be noted that when $i=j$, $(i,j)$ corresponds to a group of size 1 with only vertex with label $i$. 

\noindent The assignment rule to compute the online assignment probability of assigning resource $u$ to a group of type $v^{g}$ at step $(i,j)$ of the algorithm is defined by
{ 
\begin{align}
\frac{x_{u,v^{g}}^{*,t} \cdot \gamma}{h_{v^{g}}^{t} \cdot P_{v^{g},(i,j)}^{t} \cdot \beta_{u,(i,j)}^{t}} \label{eq:assignruleshare}
\end{align}
}
\noindent where $\beta_{u,(i,j)}^{t}$ denotes the probability that resource $u$ is available for assignment in round $t$ at step $(i,j)$ over all arrival sequences. Similarly $P_{v^{g},(i,j)}^{t}$ denotes the probability that group of type $v^{g}$ can be considered for assignment in round $t$ at step $(i,j)$ over all arrival sequences. $h_{v^g}^t$ was defined in Equation~\eqref{cons:hvg}. We use ${E}_{*,v^{g},(i,j)}^{t} \subset {\cal U}$ to denote the set of safe resources for group of type $v^{g}$ at step $(i,j)$. \\

\noindent {\ul{ \textit{Similarities and Differences to ADAPBatch: }}} The intuition behind the assignment rule for a step is similar to the one in ADAPBatch. Assignment for a group of type $v^g$ in a step is obtained by dividing the optimal assignment of round $t$ for group of type $v^{g}$ by the total number of steps where group of type $v^{g}$ can be considered~\footnote{Each group of type $v^{g}$ will be considered at $h_{v^{g}}^{t}$ steps out of the total $b^{t} \cdot b^{t}$ steps. For $\kappa = 2$, from Equation \eqref{cons:hvg} \\$h_{v^{g}}^{t} = \begin{cases} b^{t}, ~~ if ~|v^{g}| = 1,\\
  b^{t} \cdot (b^{t}-1) ~~ if ~|v^{g}| = 2~ and~v^{g} = (v,v'),\\
  \frac{b^{t} \cdot (b^{t}-1)}{2} ~~ if ~|v^{g}| = 2~ and~v^{g} = (v,v).
\end{cases}$

\noindent This is because when both vertices are of same type in the group, for example if $v^{g}=(v,v)$, then $v^g$ considered at step $(i,j)$ means that the vertex with label $i$ and the vertex with label $j$ both are $v$ and therefore steps $(i,j)$ and $(j,i)$ would be identical. On the other hand when both vertices are of different type, for example if $v^{g} =(v,v')$, then $v^{g}$ considered at step $(i,j)$ means that the vertex with label $i$ is $v$ and the vertex with label $j$ is $v'$ but $v^{g}$ considered at step $(j,i)$ means the opposite. Hence in this case the group of type $v^{g}$ will be considered at $b^{t} \cdot (b^{t} -1)$ steps across different online arrivals. Please refer to the example in the document at https://tinyurl.com/rjs524p for more clarity.
}. 

\noindent The key differences in assignment rule of ADAPShare-$\kappa$ and ADAPBatch: 
\squishlist
\item For $\kappa = 2$, since we can consider 2 vertices together (in a group) for assignment, we process the groups in $b^{t} \cdot b^{t}$ steps for ADAPShare-2. This is in comparison to $b^t$ steps in ADAPBatch. 
\item In ADAPBatch, during online processing, vertex with label $i$ in the batch will be considered for assignment only at one of $b^{t}$ steps. In ADAPShare-$\kappa$, a vertex is part of multiple groups, so it will be considered at multiple steps. Therefore, at each step, the probability of vertex being available (and as a result a group being available) needs to be recomputed based on the groups assigned at previous steps in the same round. 
\squishend 
Figure \ref{fig:adap} highlights the difference in the way the algorithms ADAPBatch and ADAPShare-$\kappa$ process the online information.\\

\noindent \textbf{Competitive Ratio for ADAPShare-2}

In this section, we provide the analysis to compute the competitive ratio for ADAPShare-2. We first find the value of $\gamma$ for which the assignment rule in Equation \eqref{eq:assignruleshare} is valid, i.e., it corresponds to a valid probability value between 0 and 1.

\begin{prop}
\label{prop:validassignrule}
The maximum value of $\gamma$ for which assignment rule in Equation \eqref{eq:assignruleshare} is valid is 0.31767. 
\end{prop}
\noindent \textbf{Proof:} Since the assignment rule always generates a positive value, the condition to be satisfied for the assignment rule to be valid is 
{
\begin{align}
& \frac{x_{u,v^{g}}^{*,t} \cdot \gamma}{h_{v^{g}}^{t} \cdot P_{v^{g},(i,j)}^{t} \cdot \beta_{u,(i,j)}^{t}} \leq 1 \label{eqn:ads} 
\end{align}
}
\noindent Using Equation \eqref{cons:hvg} in Constraint \eqref{cons:optc3} of optimization formulation in Table \ref{opt:caphigh}, we have
{ \small \begin{align}
\sum_{u} x_{u,v^{g}}^{*,t} & \leq h_{v^{g}}^{t} \cdot \prod\limits_{v\in v^{g}} (p_{v}^{t})^{n_{v,v^{g}}} \implies x_{u,v^{g}}^{*,t} & \leq h_{v^{g}}^{t} \cdot \prod\limits_{v\in v^{g}} (p_{v}^{t})^{n_{v,v^{g}}} \nonumber
\end{align}}
Substituting this in Equation~\eqref{eqn:ads} and rearranging terms, we get 
{
\begin{align}
\beta_{u,(i,j)}^{t} \geq \frac{\gamma \cdot \prod\limits_{v\in v^{g}} (p_{v}^{t})^{n_{v,v^{g}}}}{P_{v^{g},(i,j)}^{t}} ~~~ \forall t,i,j,v^{g} \label{eq:assignvalid}
\end{align}
}
\noindent By considering the probabilities with which each of the vertex of type $v \in v^{g}$ is available at step $(i,j)$, we can show that~\footnote{Please refer to https://tinyurl.com/rjs524p for the detailed proof.}, 
{
\begin{align}
& \frac{\prod\limits_{v \in v^{g}} (p_{v}^{t})^{n_{v,v^{g}}}}{P_{v^{g},(i,j)}^{t}} \leq \frac{1}{(1-\gamma)^{2}}, \forall t,i,j,v^{g} \label{eq:final}
\end{align}
}
\noindent Using Equations \eqref{eq:assignvalid} and \eqref{eq:final}, for the assignment rule to be valid it is sufficient to show that $\beta_{u,(i,j)}^{t} \geq \frac{\gamma}{(1-\gamma)^{2}}$. 

\noindent We can compute a lower bound on the value of $\beta_{u,(i,j)}^{t}$ based on assignments performed in previous steps and rounds. Specifically, using mathematical induction, we can show that $\beta_{u,(i,j)}^{t} \geq 1-\gamma$. \\

\noindent So, to find the maximum value of $\gamma$ for which the assignment rule is valid, we take $\gamma$ such that $1-\gamma = \frac{\gamma}{(1-\gamma)^{2}}$
\noindent Therefore, the possible value of $\gamma$ is the solution to the equation $\gamma = (1-\gamma)^{3}$, which is $\gamma=0.31767$.

\begin{prop}
\label{prop:adapshare}
The online algorithm ADAPShare-2 is 0.31767 competitive.
\end{prop}
\noindent \textbf{Proof:} The proof involves first showing that the ADAPShare-2 is $\gamma$ competitive. Now, as from Proposition \ref{prop:validassignrule}, the maximum value of $\gamma$ for which assignment rule is valid is 0.31767, therefore the algorithm is 0.31767 competitive. 

\noindent To show that the ADAPShare-2 is $\gamma$ competitive, we compute with respect to the optimal, the fraction of times any resource $u$ is assigned to any group of type $v^{g}$. 
The probability that the resource $u$ is assigned to a group of type $v^{g}$ in round $t$ in step $(i,j)$ is given by 
{
\begin{align}
 \frac{x_{u,v^{g}}^{*,t} \cdot \gamma}{h_{v^{g}}^{t} \cdot P_{v^{g},(i,j)}^{t} \cdot \beta_{u,(i,j)}^{t}} \cdot \beta_{u,(i,j)}^{t} \cdot P_{v^{g},(i,j)}^{t} = \frac{x_{u,v^{g}}^{*,t} \cdot \gamma}{h_{v^{g}}^{t}}\nonumber
\end{align}}
where first term in the product is the assignment rule, second term is the probability that $u$ is available and the last term is the probability that $v^{g}$ is available in round $t$ at step $(i,j)$.

\noindent As mentioned before, each group of type $v^{g}$ will be considered for assignment at a total of $h_{v^{g}}^{t}$ steps. Therefore, the expected number of times a resource $u$ is assigned to a group of type $v^{g}$ in round $t$ is given by 
$h_{v^{g}}^{t} \cdot \frac{x_{u,v^{g}}^{*,t} \cdot \gamma}{h_{v^{g}}^{t}} = x_{u,v^{g}}^{*,t} \cdot \gamma $, i.e., in online case each resource $u$ is matched to group of type $v^{g}$ with probability equal to $x_{u,v^{g}}^{*,t}\cdot \gamma$. 

\noindent Therefore, ADAPShare-2 is $\gamma$ competitive. $\blacksquare$


\begin{corollary-1}
\label{cor:adapshare}
The online algorithm ADAPShare-$\kappa$ (generalization of ADAPShare-2 for any value of $\kappa$) is $\gamma$ competitive where the value of $\gamma$ is the solution to the equation $\gamma = (1-\gamma)^{\kappa+1}$.
\end{corollary-1}
\noindent {\bf Proof Sketch:} The proof
is along the same lines as the proof for Proposition \ref{prop:adapshare}. In the Equation \eqref{eq:final}, instead of $(1-\gamma)^2$, we will have $(1-\gamma)^{\kappa}$. Therefore, the value of $\gamma$ for which assignment rule is valid is the solution to the Equation $\gamma = (1-\gamma)^{\kappa+1}$. $\blacksquare$ \\
 
\noindent \textbf{Hardness Result for Non-Adaptive Algorithms:} 

\noindent Dickerson {\em et.al.}~\cite{dickerson2017allocation} prove that no non-adaptive algorithm based on LPSequential can achieve a competitive ratio of more than $\frac{1}{2} + o(1)$ in OM-RR-KAD model. The analysis can be easily extended for the batch arrival case when $\kappa = 1$. As unit-capacity batch arrival is a special case of multi-capacity OPERA model with all $w_{u,v^{g}}^{t}=0, ~if~ |v^{g}| \geq 2$, therefore, no non-adaptive algorithm based on LPShare can achieve a competitive ratio of more than $\frac{1}{2} + o(1)$ for OPERA model.\\ 

\noindent \textbf{Discussion:}We now provide the justifications for the choices made in the modelling and analysis in section \ref{sect:extension} and \ref{sect:model}. (1) We assume that there are $b^t$ arrivals in round t and $b^t$ is known in advance. However, this is not at all a strong assumption because by considering a null type vertex in ${\cal V}$ and $p_{\phi}^t$ as the probability of null vertex, $b^t$ can be used to denote the maximum number of arrivals in round $t$. (2) For theoretical analysis of the solution quality, we ignore the computational complexity of generating exponential number of groups in OPERA model. For practical purposes, the algorithms provided in \cite{alonso2017demand} can be used to heuristically prune the exponential set and generate the feasible groups efficiently. The pruned set of groups is used by both offline and online algorithms. This is because, if the offline optimal algorithm can generate the groups, as the type of vertices are known in advance (through the known distribution), the online algorithm can also use those groups.
\section{Experiments}
In this section, we compare the following five approaches on the empirical competitive ratio metric: 
\squishlist
\item {\bf Greedy} - Runs an integer optimization at each timestep (based on the current information) to assign the requests/groups to the available offline resources~\footnote{Equivalent to the myopic approaches used in practice~\cite{alonso2017demand,lowalekarVJ19}}. 
\item {\bf Random} - Shuffles available requests/groups randomly and then assigns each request/group randomly to an available offline resource. 
\item {\bf Alg-OPERA-1 } - Algorithm based on the offline optimal LP where match for any available resource $u$ to a vertex or group is performed by looking at the value of $\frac{x^{*,t}_{u,v^{g}}}{q_{v^{g}}^{t}}$~\footnote{We provide heuristics, which are close to ADAPShare-$\kappa$, as computing $\beta$ exactly is not always simple and may require large number of simulations. We observed that even though these heuristics are non-adaptive, they can achieve empirical competitive ratio higher than the theoretical competitive ratio of ADAPShare-$\kappa$.}. 
\item {\bf Alg-OPERA-2} - Another algorithm based on the offline optimal LP where match for any available resource $u$ to a vertex or a group is performed by looking at the value of $\frac{x^{*,t}_{u,v^{g}}}{\sum_{u} x^{*,t}_{u,v^{g}}}$.
\item {\bf $\epsilon$-Greedy} - With probability $\epsilon$, greedy algorithm is executed and with probability $1-\epsilon$, Alg-OPERA-1 algorithm is executed. 
\squishend
The goal of the experiments is to show that the algorithms which use guidance from the offline optimal LP, outperform the myopic approaches~\footnote{Currently used in practice for multi-capacity resources~\cite{alonso2017demand,lowalekarVJ19}}, which do not consider future information. All the values in the results are computed by taking an average over 10 instances and each instance is run 100 times. 

\noindent \textbf{Synthetic Dataset:} We first present the results on a synthetic dataset. We use 200 timesteps/rounds and generate the unavailability (or time occupied serving requests) time ($c_{u,v}^{t}$ or $c_{u,v^{g}}^{t}$) for each resource and vertex/group pair randomly between 1 and 60. Weights received (revenue) are generated based on revenue model used by taxi companies -- base revenue + $0.5 \cdot c_{u,v}^{t}~ or~ c_{u,v^{g}}^{t}$. The probability of arrival of each vertex type at each round ($p_{v}^{t}$) is also generated randomly. The test instances are generated by sampling the online vertices from the generated $p_{v}^{t}$ values. We vary the batch size and capacity and present the representative results.

\begin{figure}
  \centering
    \subfloat[]{\label{fig:figcap}\includegraphics[width=0.45\textwidth,height=2.2in]{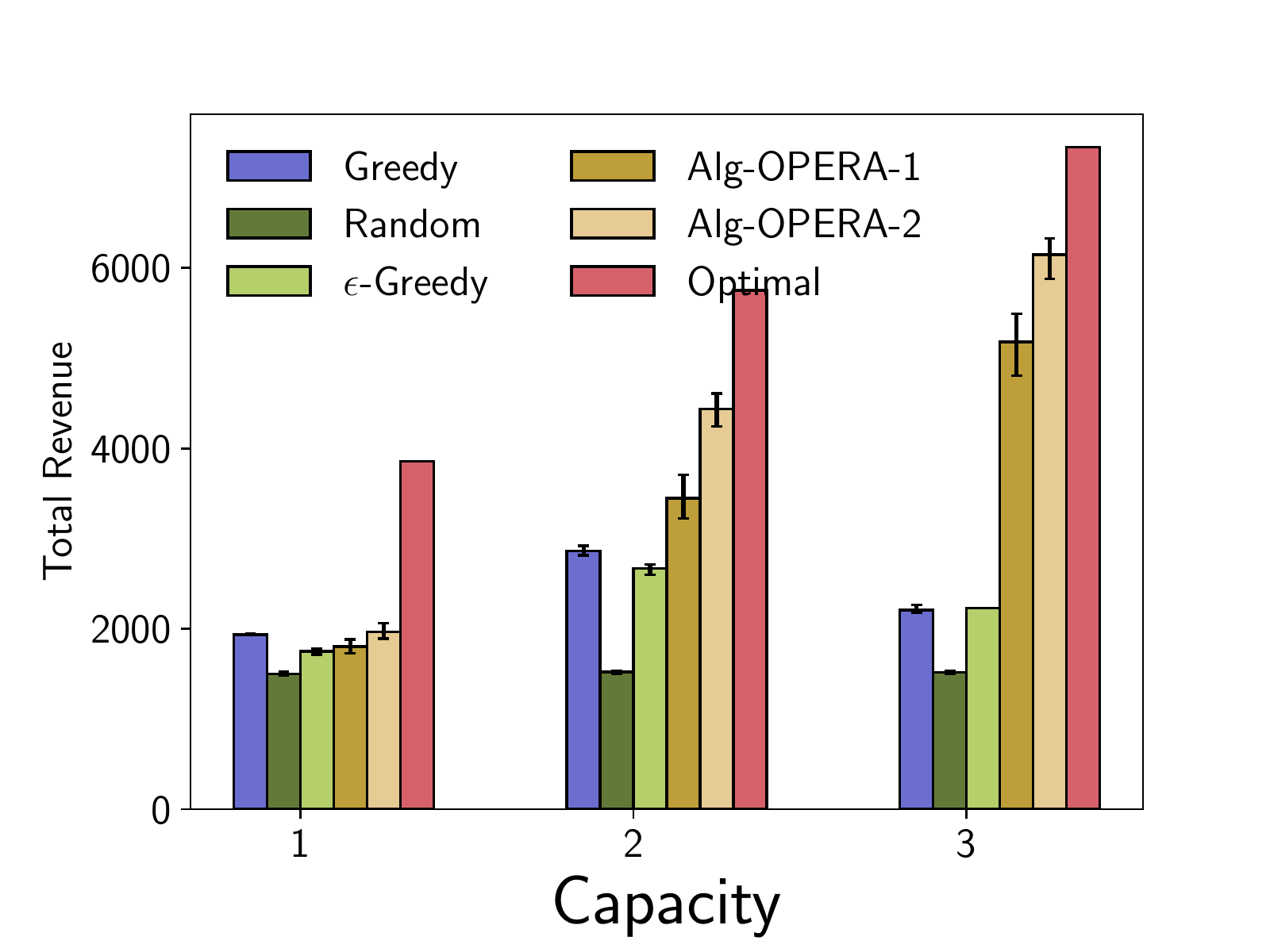}}
    \subfloat[]{\label{fig:bsizecap2}\includegraphics[width=0.45\textwidth,height=2.2in]{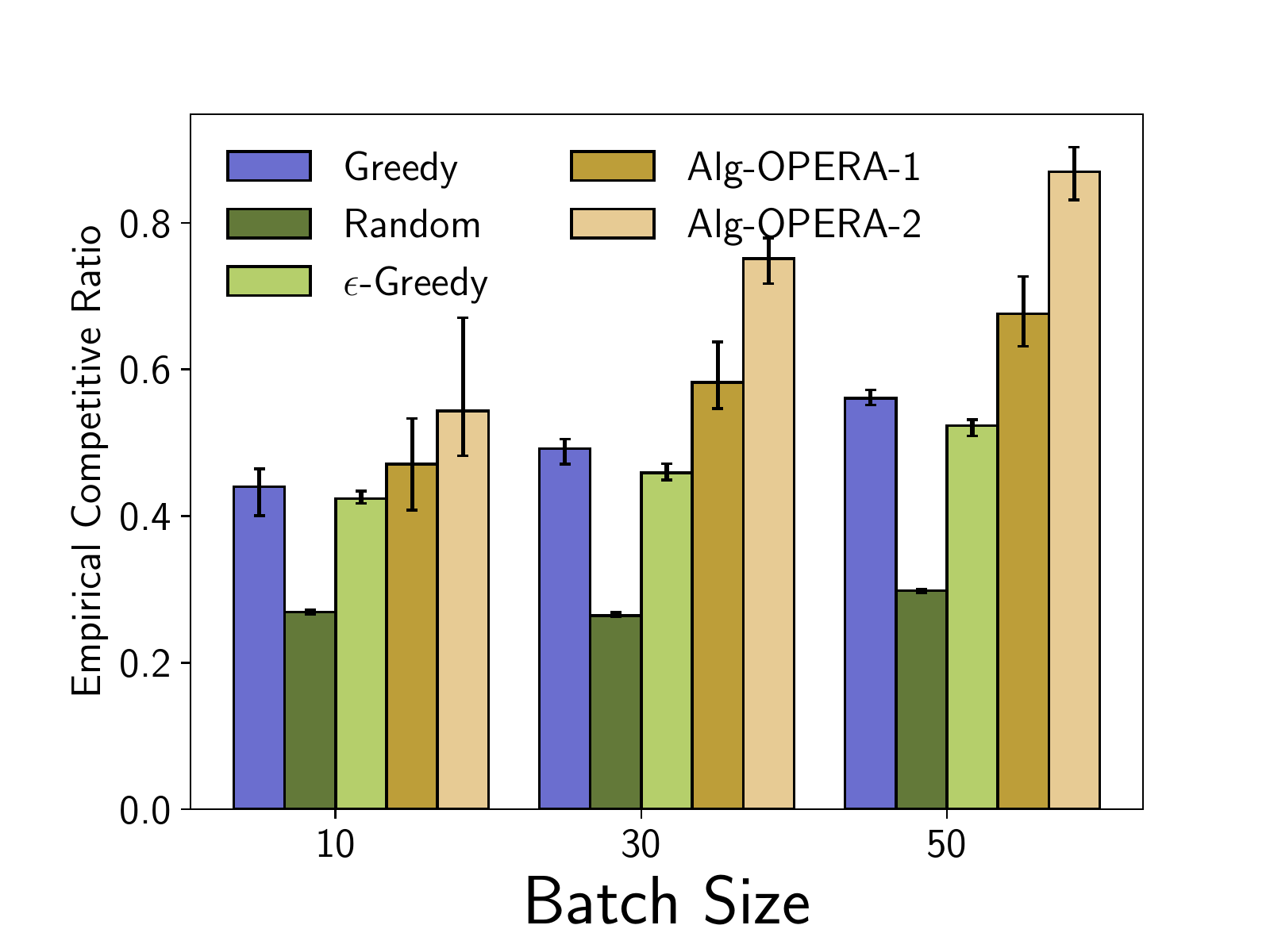}}
  \caption{{{$|{\cal U}|$ = 10 , $|{\cal V}|$ =10, $T=200$ (a) Varying $\kappa$ (b) $\kappa =2$}}}
  \label{fig:figresults}
\end{figure}

\noindent Figure \ref{fig:figcap} shows the total revenue obtained by different algorithms for different values of capacity $\kappa$. The key observations are: \\
\noindent (1) Our online approaches (Alg-OPERA-1 and Alg-OPERA-2) outperform other algorithms, with Alg-OPERA-2 performing better than Alg-OPERA-1 on all the instances. \\
\noindent (2) The performance of greedy algorithm decreases with the increase in capacity. Higher capacity provides more opportunity to serve requests at each timestep. Due to its myopic nature, greedy algorithm serves more requests initially, keeping the resources occupied for a longer time. On the other hand Alg-OPERA-1 and Alg-OPERA-2, based on the guidance provided by the offline optimal LP, ignore some requests/groups which have higher $c_{u,v^{g}}^{t}$ value, to serve more requests at future timesteps. \\
\noindent (3) Figure \ref{fig:bsizecap2} shows the empirical value of competitive ratio for different batch sizes. For these experiments, we take the identical value of batch size for all the timesteps. Higher batch size for multi-capacity resources provides an opportunity to group more requests. Therefore, as the batch size increases Alg-OPERA-1 and Alg-OPERA-2 show an improvement in performance. 

\begin{figure}
  \centering
    \subfloat[]{\label{fig:nycap}\includegraphics[width=0.45\textwidth,height=2.2in]{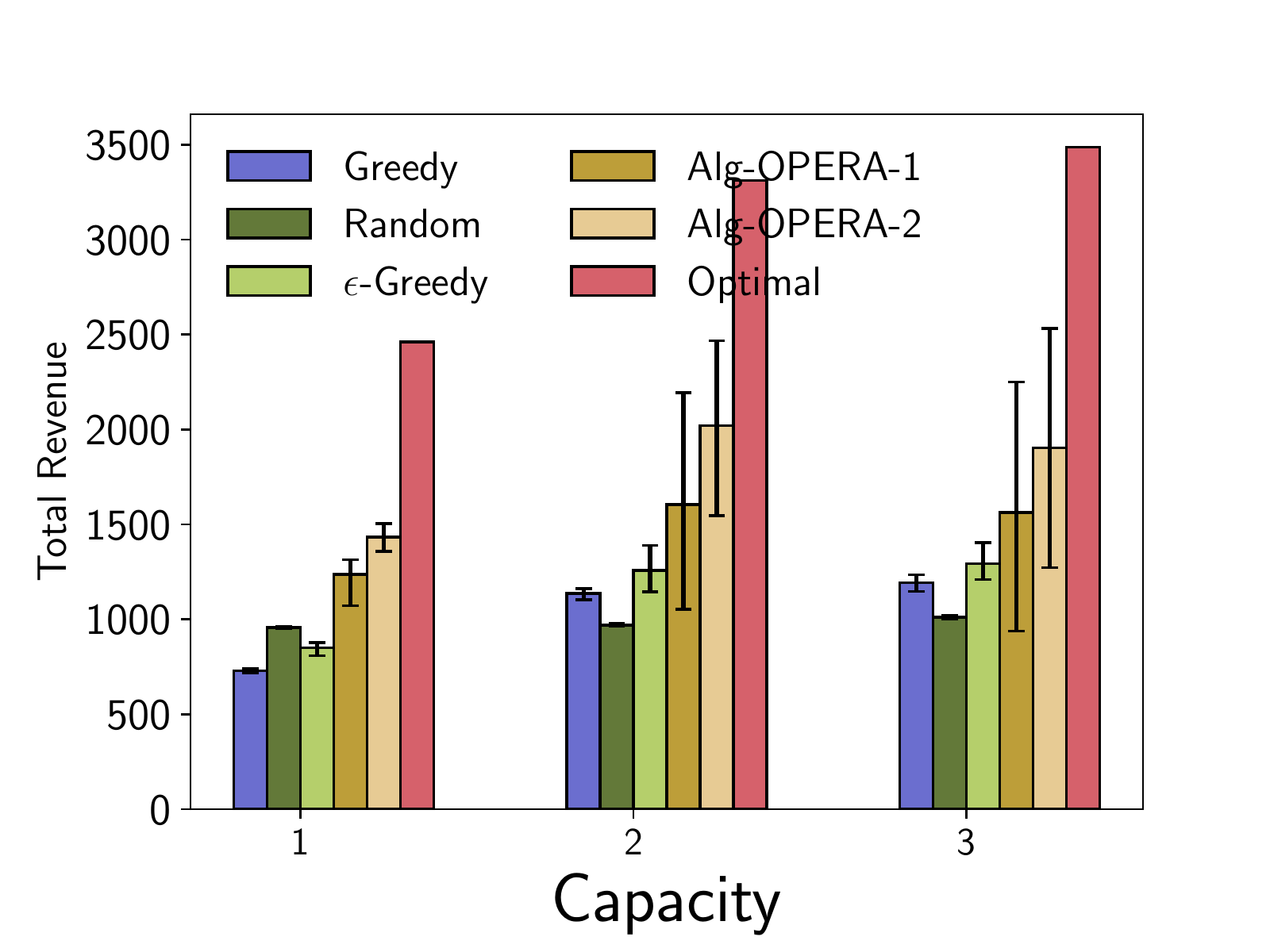}}
    \subfloat[]{\label{fig:nycap1}\includegraphics[width=0.45\textwidth,height=2.2in]{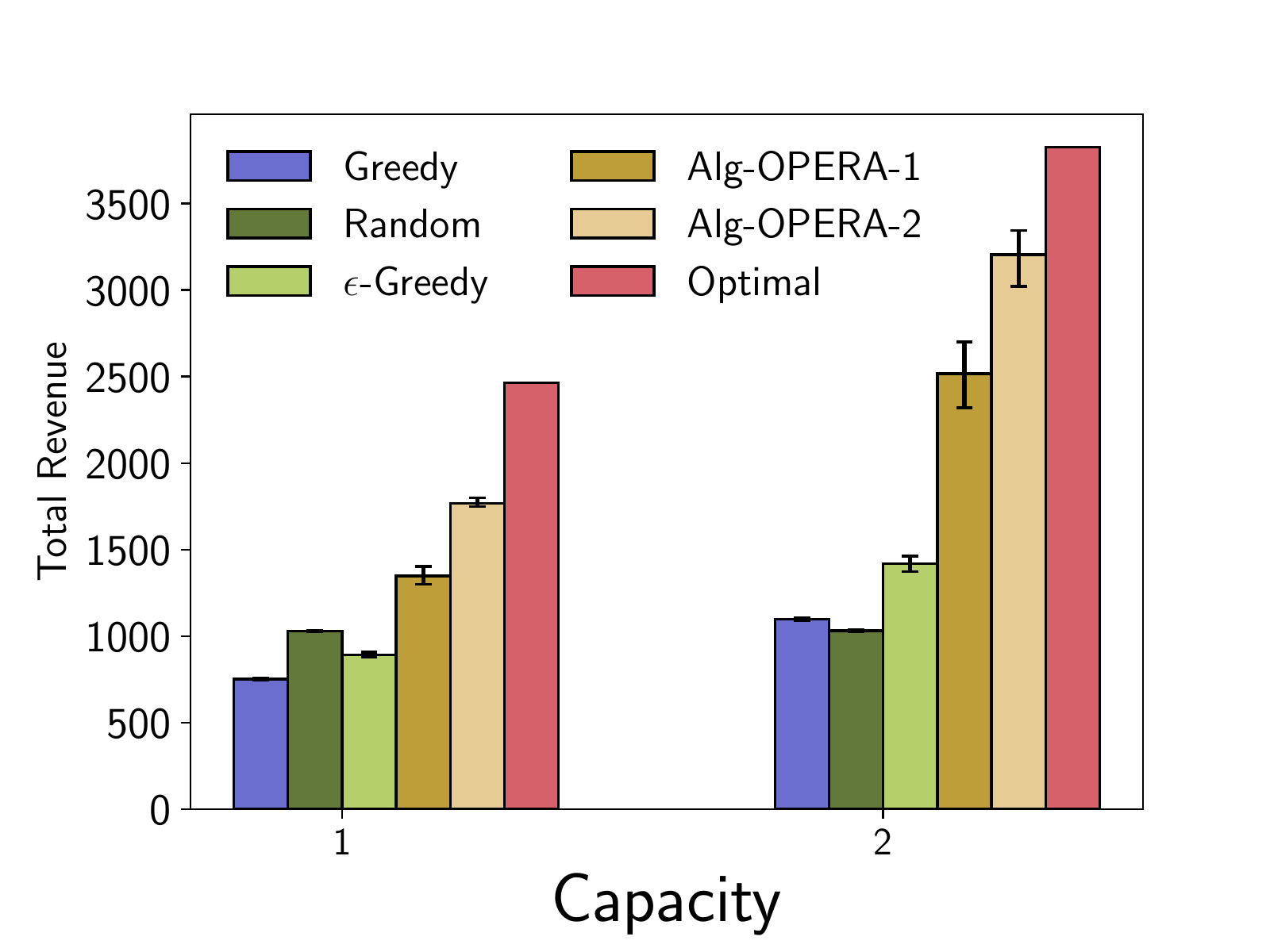}}
  \caption{{ $|{\cal U}|$ = 30 , $|{\cal V}|$ =121, $T=240$, Real Dataset (a) 12am (b) 8am }}
  \label{fig:figresults-real}
\end{figure}

\textbf{Real World Dataset:} We used the New York Yellow Taxi dataset which contains the records of trips in Manhattan city. We divided the map of the city into a grid of squares, each 4 by 4 km, which resulted in a total of 11 squares.
Therefore, there can be 121 different types of requests, i.e., $|{\cal V}| = 121$ (origin-destination pairs). We experimented by taking real trips from the taxi dataset. We take the data across 10 days to compute the $p_{v}^{t}$ values and the average number of requests at each round/timestep, i.e., average value of $b^{t}$. We run the offline optimal LP with these values and get a solution. The online algorithms are tested on actual instances (10 days) which are different from the ones we used for computing the parameter values. Therefore, the actual batch size $b^{t}$ can be different from the value used by an offline optimal solution. The taxis are initialized at random locations and since we are testing the last mile scenario after serving the trips, they come back to their starting location. We observe a high variance in the performance of our algorithms on this dataset during night time (Figure \ref{fig:nycap}). This is because the distribution of requests during night have high variance across days. During the day, the variance in distribution of requests is low, and as a result our algorithms also show low variance. On an average , Alg-OPERA-1 and Alg-OPERA-2 outperform other algorithms on this dataset as well. These results indicate that the algorithms which use the guidance from offline optimal solution can consider the future effects of current matches and as a result provide better performance.

\noindent We would like to highlight that, to ensure that the theoretical bound on the competitive ratio holds empirically, correct estimates of probability values ($p^{t}_{v},\beta$) are required, which requires running multiple simulations. It is possible to create scenarios, where a high number of simulations are required to get the correct estimates (e.g., when all the $p_{v}^{t}$ values are very small and ${\cal V}$ is large.). In such cases, empirical competitive ratio measured over low number of simulations, will be a wrong indicator. We would also like to mention that, it is possible to synthetically create unrealistic scenarios where Greedy algorithm can achieve close to optimal value (essentially having a revenue model such that the difference between one long trip and multiple short trips is almost negligible, so myopic decisions do not hurt) and can perform better than the LP based approaches.

\section{Conclusion}
In this paper, we make a fundamental contribution of providing competitive ratios for the challenging online multi-capacity ridesharing problems -- where resources or vehicles retain their type after serving the requests and rejoining the system -- with batch arrival of requests. We demonstrate empirically on real and synthetic datasets that our online heuristics based on offline optimal LP perform well in practice, as compared to the myopic approaches.

\section{Acknowledgements}
This work was partially supported by the Singapore National Research Foundation through the Singapore-MIT Alliance for Research and Technology (SMART) Centre for Future Urban Mobility (FM). We thank Sanket Shah, Susobhan Ghosh and Tanvi Verma for providing valuable comments which greatly improved the paper.
\section*{References}
\appendix
\bibliography{rideref}

\begin{table}[h]
\center
        \begin{tabular}{|r|}
        \hline

        \begin{minipage}{0.95\textwidth}
                \vspace{0.05in}
\textbf{LPBatch:}
{
\begingroup
\addtolength{\jot}{-2pt}
{\begin{align}
\max & \sum_{t \in T} \sum_{u \in {\cal U}} \sum_{v \in {\cal V}} w_{u,v}^{t} \cdot x_{u,v}^{t} \nonumber \\
s.t. \quad & \sum_{u \in {\cal U}} x_{u,v}^{t} \leq q_{v}^{t} ::: \forall v,t \label{cons:optb1} \\
& \sum_{t' < t} \sum_{v' \in {\cal V}} x_{u,v'}^{t'} \cdot Pr[c_{u,v'}^{t'} > t-t'] + \sum_{v \in {\cal V}} x_{u,v}^{t} \leq 1 \label{cons:optb2},\forall  u,t \\
& 0 \leq x_{u,v}^{t} \leq 1 ::: \forall u,v,t 
\end{align}}
\endgroup }
\end{minipage} \\
        \hline
        \end{tabular}
        \caption{Optimization Formulation - Unit Capacity Batch Arrival}
        \label{opt:batch}
        \end{table}

\section{Proof of Proposition 1}
\begin{prop}
The optimal value of the LP presented in Table \ref{opt:batch} is a valid upper bound on the offline optimal value. 	
\end{prop}

\noindent \textbf{Proof:} 
To show that the optimal value of the LP provides a valid upper bound on the offline optimal value, we prove that the expected value of the optimal matching is less than or equal to the optimal value of the LP. We have two distributions, first one corresponding to the arrival of vertices ($p_{v}^{t}$) and another corresponding to the number of rounds of unavailability ($c_{u,v}^{t}$). To prove that the optimal value of LP provides a valid upper bound in the presence of both distributions, similar to earlier works \cite{karthik2019}, we need to make an assumption that the optimal assignment only depends on the arrival distribution and is independent of the value of $c_{u,v}^{t}$~\footnote{Instead of this, we can also assume that $c_{u,v}^{t}$ is a known constant which is a weaker assumption than assuming that the optimal assignment is independent of distribution. The proof is similar for both assumptions.}.

Consider a realization of arrivals denoted by sequence $a$. Let $m_{v}^{t}(a)$ denotes the number of vertices of type $v$ arriving in round $t$	 for arrival sequence $a$. Similarly, consider a realization of the number of rounds of unavailability denoted by sequence $r$, where $c_{u,v}^{t}(r)$ denotes the number of rounds for which resource $u$ becomes unavailable on matching with vertex of type $v$ in sequence $r$. Let $\delta(c_{u,v}^{t'}(r) > t-t')$ is an indicator variable denoting that the number of rounds for which resource $u$ becomes unavailable on being assigned to vertex of type $v$ in round $t'$ is greater than $t-t'$ rounds in sequence $r$. Now, for any arrival sequence $a$ and any realization of the number of rounds of unavailability $r$, the offline solution can be computed by solving the optimization program in Table \ref{opt:batchint}~\footnote{We index each of $m_{v}^{t}$, $x_{u,v}^{t}$  by $a$ and $c_{u,v}^{t}$ by $r$ to denote the instance for sequence $a$ and $r$.}. 

As mentioned before, we make an assumption that the optimal assignment is independent of the realization of the number of rounds of unavailability, i.e., it only depends on $a$ and not on $r$. Therefore, we use $x^{*}(a)$ to denote the optimal solution of the formulation in Table \ref{opt:batchint} for sequence $a$ and $r$. The expected number of times edge $(u,v)$ is matched at $t$ is given by $\sum_{a} x^{*,t}_{u,v}(a) \cdot P(a)$. To prove that, the LP in Table \ref{opt:batch} is a valid upper bound on the offline optimal value, we show that $\forall u,v,t ~~\sum_{a}x^{*,t}_{u,v}(a)\cdot P(a)$ is a feasible solution to the LP.

The optimization in Table \ref{opt:batchint} is solved for each sequence $a$ and $r$. As $x^{*}(a)$ denotes the optimal assignment for the sequence $a$ and $r$, therefore, $x^{*}(a)$ satisfies the constraints of the formulation in Table \ref{opt:batchint}. Hence, we get

\begin{align}
& \sum_{u \in {\cal U}} x^{*,t}_{u,v}(a) \leq m_{v}^{t}(a) ::: \forall v,t \label{cons:opt21_a}\\
& \sum_{t' < t} \sum_{v' \in {\cal V}} x^{*,t'}_{u,v'}(a) \cdot Pr(c_{u,v'}^{t'}(r) > t-t') + \sum_{v \in {\cal V}} x^{*,t}_{u,v}(a) \leq 1 \label{cons:opt22_a}::: \forall  u,t 
\end{align}

Multiplying Equations \eqref{cons:opt21_a} and \eqref{cons:opt22_a} by $P(a) \cdot P(r)$~\footnote{$a$ and $r$ are independently drawn from the distributions.}, probability of sequence $a$ and $r$, and performing summation over $a$ and $r$, we get

\begin{align}
& \sum_{u \in {\cal U}} \sum_{a} \sum_{r} x^{*,t}_{u,v}(a) \cdot P(a) \cdot P(r) \leq \sum_{a} \sum_{r} m_{v}^{t}(a) \cdot P(a) \cdot P(r) ::: \forall v,t \label{cons:opt21_a_1}\\
& \sum_{t' < t} \sum_{v' \in  {\cal V}} \sum_{a} \sum_{r} x^{*,t'}_{u,v'}(a) \cdot \delta(c_{u,v'}^{t'}(r) > t-t') \cdot P(a) \cdot P(r) \nonumber \\
& \hspace{1in}+ \sum_{v \in {\cal V}} \sum_{a} \sum_{r}x^{*,t}_{u,v}(a) \cdot P(a) \cdot P(r) \leq \sum_{a} \sum_{r} 1 \cdot P(a) \cdot P(r) \label{cons:opt22_a_1}::: \forall  u,t 
\end{align}
$\sum_{a} \sum_{r}1 \cdot P(a) \cdot P(r) = 1$ and $\sum_{a} \sum_{r}m_{v}^{t}(a) \cdot P(a) \cdot P(r) = \sum_{a} m_{v}^{t} \cdot P(a)$ denotes the expected number of vertices of type $v$ arriving in round $t$. Therefore, 

$$\sum_{a}m_{v}^{t} \cdot P(a) = q_{v}^{t}$$.

As $x^{*}$ is independent of $r$, therefore, 

$$\sum_{a} \sum_{r} x^{*,t}_{u,v}(a) \cdot  P(a) \cdot P(r) = \sum_{a} x^{*,t}_{u,v}(a) \cdot P(a)$$  

On substituting these values, we get,

\begin{align}
& \sum_{u \in {\cal U}} \sum_{a} x^{*,t}_{u,v}(a) \cdot P(a) \leq q_{v}^{t} ::: \forall v,t \label{cons:opt21_a_11}\\
& \sum_{t' < t} \sum_{v' \in  {\cal V}} \sum_{a} x^{*,t'}_{u,v'}(a) \cdot  P(a)  \cdot  \sum_{r} \delta(c_{u,v'}^{t'}(r) > t-t') \cdot P(r) + \sum_{v \in {\cal V}} \sum_{a} x^{*,t}_{u,v}(a) \cdot P(a) \leq 1 \label{cons:opt22_a_12}::: \forall  u,t 
\end{align}

As, 

$$\sum_{r} \delta(c_{u,v'}^{t'}(r) > t-t') \cdot P(r) = Pr(c_{u,v'}^{t'} > t-t') $$

Therefore, the Equations \eqref{cons:opt21_a_11} and \eqref{cons:opt22_a_12} are same as the constraints of the optimization formulation in Table \ref{opt:batch} with $x_{u,v}^{t} = \sum_{a} x^{*,t}_{u,v}(a) \cdot P(a)$. Therefore, $\sum_{a} x^{*}(a) \cdot P(a)$ is a feasible solution to the LP in Table \ref{opt:batch}. As the value of the optimal solution is greater than or equal to the value of any feasible solution, hence it is proved that the optimal value of LP provides a valid upper bound on the offline optimal value. 


\begin{table}[t]

\center
        \begin{tabular}{|r|}
        \hline

        \begin{minipage}{0.95\textwidth}
                \vspace{0.05in}
{
\begingroup
\addtolength{\jot}{-2pt}
\begin{align}
\max & \sum_{t \in T} \sum_{u \in {\cal U}}\sum_{v \in {\cal V}} w_{u,v}^{t} \cdot x_{u,v}^{t}(a) \\
s.t. \quad & \sum_{u \in {\cal U}} x_{u,v}^{t}(a) \leq m_{v}^{t}(a) ::: \forall v,t \label{cons:optint21}\\
& \sum_{t' < t} \sum_{v' \in  {\cal V}} x_{u,v'}^{t'}(a) \cdot \delta(c_{u,v'}^{t'}(r) > t-t') + \sum_{v \in {\cal V}} x_{u,v}^{t}(a) \leq 1 \label{cons:optint22}::: \forall  u,t \\
& x_{u,v}^{t}(a) \in \{0, 1\} 
\end{align}
\endgroup }
\end{minipage} \\
        \hline
        \end{tabular}
        \caption{ Optimization Formulation - Batch Arrival - For a fixed sequence $a$ and $r$ }
        \label{opt:batchint}
        \end{table}

\section{Proof of Proposition 2}
\begin{prop}
The online algorithm ADAPBatch is $\frac{1}{2}$ competitive.
\end{prop}

\noindent\textbf{Proof:} The proof proceeds in two steps:
\begin{enumerate}
\item We first prove that the maximum value of $\gamma$ for which the assignment rule of ADAPBatch is valid is $\frac{1}{2}$.
\item Next, we prove that the online algorithm is $\gamma$ competitive. As the maximum value of $\gamma$ for which the assignment rule is valid is $\frac{1}{2}$, therefore, the online algorithm is $\frac{1}{2}$ competitive.
\end{enumerate}

\textbf{Computing Maximum value of $\gamma$ for which the assignment rule is valid:}
Since $x^{*,t}_{u,v} \geq 0 , \forall u,v,t$, the assignment rule will always generate a positive value. Therefore, the only condition which should be satisfied for the batch assignment rule to be valid is, 
\begin{align}
\frac{x^{*,t}_{u,v} \cdot \gamma}{b^{t} \cdot p_{v}^{t} \cdot \beta_{u,i}^{t}} \leq 1 ::: \forall u,v,i,t\label{eq:assignrulebatch}
\end{align}

Using the expression $q_{v}^{t}=b^{t}\cdot p_{v}^{t}$ in Constraint \eqref{cons:optb1} of the optimization formulation in Table \ref{opt:batch}, we have, 
$$ \sum_{u} x^{*,t}_{u,v} \leq b^{t}\cdot p_{v}^{t} \implies x^{*,t}_{u,v} \leq b^{t}\cdot p_{v}^{t} $$

Substituting this in Equation \eqref{eq:assignrulebatch} and rearranging terms, we get
\begin{align} 
\beta_{u,i}^{t} & \geq \gamma, \forall u,i,t \label{eq:assignvalideq}
\end{align}
 
Therefore, we focus on finding the value of $\beta_{u,i}^{t}, \forall u,i,t$. We use mathematical induction to prove that $\beta_{u,i}^{t} \geq 1-\gamma, \forall u,i,t$. 

$\beta_{u,i}^{t}$ denotes the probability that the resource $u$ is safe in round $t$ while considering the $i^{th}$ vertex. Initially, as all the resources are available, therefore,

$$\beta_{u,1}^{1} = 1 \geq 1 - \gamma , \forall u$$

Please note that $\beta_{u,i}^{t} \geq \beta_{u,i-1}^{t}, \forall i > 0$, therefore, we only show the computation for the value of $\beta_{u,b^{1}}^{1}$. 

$\beta_{u,i}^{t}$ for any $i$ and $t$ is computed by computing the probability that $u$ is assigned to any $v$ before round $t$ step $i$ such that it has not rejoined the system yet. 

$$\beta_{u,b^{1}}^{1} = 1 - \sum_{v} \sum_{i=1}^{b^{1}-1}\beta_{u,i}^{1} \cdot p_{v}^{1} \cdot \frac{x^{*,1}_{u,v} \cdot \gamma}{b^{1} \cdot p_{v}^{1} \cdot \beta_{u,i}^{1}}= 1 - (b^{1}-1) \cdot \sum_{v} \frac{x^{*,1}_{u,v}\cdot \gamma}{b^{1}} \geq 1-\gamma (As~ b^{1} ~\geq 1)$$

As it is valid for $t=1$, we prove it by induction. Assume that it is valid for all $t' < t$, i.e., in online case the edge $(u,v)$ is matched  $x^{*,t}_{u,v} \cdot \gamma$ times in round $t$. Therefore,

$$\beta_{u,1}^{t} = 1 - \sum_{v} \sum_{t' < t} (x^{*,t'}_{u,v}\cdot \gamma \cdot Pr(c_{u,v}^{t'} > t - t'))$$

From Equation \eqref{cons:optb2} of the optimization program in Table \ref{opt:batch}, we have 

$$\sum_{v} \sum_{t' < t} (x^{*,t'}_{u,v} \cdot \gamma \cdot Pr(c_{u,v}^{t'} > t - t')) \leq \gamma - \sum_{v} x^{*,t}_{u,v} \cdot \gamma $$

Multiplying both sides of the above equation by -1, we get

$$-1 \cdot \sum_{v} \sum_{t' < t} (x^{*,t'}_{u,v} \cdot \gamma \cdot Pr(c_{u,v}^{t'} > t - t')) \geq -1 \cdot (\gamma - \sum_{v} x^{*,t}_{u,v} \cdot \gamma)$$

Adding 1 on both sides in above equation, we get

$$1 - \sum_{v} \sum_{t' < t} (x^{*,t'}_{u,v} \cdot \gamma \cdot Pr(c_{u,v}^{t'} > t - t')) \geq 1 - \gamma + \sum_{v} x^{*,t}_{u,v} \cdot \gamma$$
$$1 - \sum_{v} \sum_{t' < t} (x^{*,t'}_{u,v} \cdot \gamma \cdot Pr(c_{u,v}^{t'} > t - t')) \geq 1 - \gamma + \sum_{v} x^{*,t}_{u,v} \cdot \gamma$$

Therefore, 
$$\beta_{u,1}^{t} \geq   1 -\gamma + \sum_{v} x^{*,t}_{u,v} \cdot \gamma \geq 1-\gamma$$

Similarly, we compute $\beta_{u,b^{t}}^{t}$, by computing the probability that $u$ is assigned to any $v$ before round $t$ step $i$ such that it has not rejoined the system yet.

$$\beta_{u,b^{t}}^{t} = 1 - \sum_{v} \sum_{t' < t} (x^{*,t'}_{u,v} \cdot \gamma \cdot Pr(c_{u,v}^{t'} > t - t')) - \sum_{v} \sum_{i=1}^{b^{t}-1} \beta_{u,i}^{t} \cdot p_{v}^{t} \cdot \frac{x^{*,t}_{u,v} \cdot \gamma}{b^{t} \cdot p_{v}^{t} \cdot \beta_{u,i}^{t}} \geq 1 - \gamma + \sum_{v} x^{*,t}_{u,v} \cdot \gamma - \sum_{v} (b^{t}-1) \cdot \frac{x^{*,t}_{u,v} \cdot \gamma }{b^{t}} \geq 1-\gamma$$

Therefore, $\beta_{u,i}^{t} \geq 1-\gamma, \forall u,i,t$. From Equation \eqref{eq:assignvalideq}, for assignment rule to be valid $\beta_{u,i}^{t} \geq \gamma$. Therefore, the maximum possible value of $\gamma$ is the solution of equation $1-\gamma = \gamma \implies \gamma= \frac{1}{2}$.

\textbf{Online Algorithm ADAPBatch is $\gamma$ competitive:} Let $M_{u,v,i}^{t}$ is an indicator variable denoting that the resource $u$ is tried for assignment to vertex of type $v$ in round $t$ while processing $i^{th}$ vertex of the batch. Let $N_{u,i}^{t}$ is an indicator variable denoting that the resource $u$ is available in round $t$ while processing $i^{th}$ vertex of the batch and $O_{v,i}^{t}$ denotes that the vertex of type $v$ is processed as $i^{th}$ vertex in round $t$.
-
Therefore, probability that the resource $u$ is assigned to the vertex of type $v$ while processing $i^{th}$ vertex of the batch in round $t$ is given by 

$$P[M_{u,v,i}^{t}=1] \cdot P[N_{u,i}^{t} = 1] \cdot P[O_{v,i}^{t} = 1] = \frac{x^{*,t}_{u,v} \cdot \gamma}{b^{t} \cdot p_{v}^{t} \cdot \beta_{u,i}^{t}} \cdot \beta_{u,i}^{t} \cdot p_{v}^{t} = \frac{x^{*,t}_{u,v} \cdot \gamma}{b^{t}}$$

Expected Number of times vertex of type $v$ is matched to $u$ in round $t$ = $$b^{t} \cdot \frac{x^{*,t}_{u,v} \cdot \gamma}{b^{t}} = x^{*,t}_{u,v} \cdot \gamma$$

i.e., in online case the edge $(u,v)$ is matched $x^{*,t}_{u,v} \cdot \gamma$ times in round $t$.

As each edge is made with the probability $x^{*,t}_{u,v} \cdot \gamma$ and the maximum value of $\gamma$ for which the assignment rule is valid is $\frac{1}{2}$, therefore, using Lemma~\ref{lemmakarthik} the competitive ratio of the algorithm is $\gamma=\frac{1}{2}$. 

\begin{lemm}
\label{lemmakarthik}
\cite{karthik2019}  Let $x^{*}$ denote the optimal solution to LP in Table \ref{opt:batch}. Suppose we have that for every edge $(u,v)$ at any round $t$, Pr[$(u,v)$ is included in the matching] $\geq \gamma \cdot x^{*,t}_{u,v}$ then the competitive ratio is at least $\gamma$.
\end{lemm}


\section{Expected Number of times group of type $v^g$ can be formed in round $t$}

Let $q_{v^g}^{t}$ denote the expected number of times group of type $v^{g}$ can be formed in round $t$. As each of the $b^{t}$ vertices are sampled independently from a categorical distribution $p_{v}^{t},~\forall v$, we can consider it as having $b^{t}$ trials and use $Y_{v,i}$ as an indicator variable denoting that $v$ is sampled in the $i^{th}$ trial (as the  $i^{th}$ vertex or not, out of $b^{t}$ vertices). 
Please note that $Y_{v,i} \cdot Y_{v',i} = 0, v\neq v'$ and if $i\neq j$, $Y_{v,i}$ and $Y_{v',j}$ are independent. 

We first consider a simple case, where $v^{g} = (v,v'), v\neq v'$. Therefore, if any of two different vertices are of type $v$ and $v'$ then we can form the group of type $v^{g}$. So, we have
\begin{align}
q_{v^{g}}^{t} = E\Big[\sum_{i,j;i\neq j} Y_{v,i} \cdot Y_{v',j}\Big] \nonumber
\end{align}

By linearity of expectation, we get

\begin{align}
q_{v^{g}}^{t} = \sum_{i,j;i\neq j} E[Y_{v,i} \cdot Y_{v',j}]
\end{align}

As each vertex is independently sampled, the event of sampling $i^{th}$ vertex and $j^{th}$ vertex are independent of each other. Therefore, 
\begin{align}
q_{v^{g}}^{t} = \sum_{i,j;i\neq j} E[Y_{v,i}] \cdot E[Y_{v',j}]\nonumber \\
E[Y_{v,i}^{t}] = p_{v}^{t} \nonumber \\
E[Y_{v',j}^{t}] = p_{v'}^{t}  \nonumber \\
q_{v^{g}}^{t} = \sum_{i,j;i\neq j} p_{v}^{t} \cdot p_{v'}^{t} \nonumber \\
q_{v^{g}}^{t} = b^{t} \cdot (b^{t}-1) p_{v}^{t} \cdot p_{v'}^{t} \nonumber
\end{align}

Similarly if we have $v^{g} = (v,v)$, we can form the group if any of the two vertices are of type $v$ 
\begin{align}
q_{v^{g}}^{t} = E\Big[\sum_{i,j;i <  j} Y_{v,i} \cdot Y_{v,j}\Big] \nonumber
\end{align}
By linearity of expectation
\begin{align}
q_{v^{g}}^{t} = \sum_{i,j;i < j} E[Y_{v,i} \cdot Y_{v,j}]\nonumber
\end{align}

As each vertex is independently sampled, event of sampling $i^{th}$ vertex and $j^{th}$ vertex are independent. Therefore, 

\begin{align}
q_{v^{g}}^{t} = \sum_{i,j;i < j} E[Y_{v,i}] \cdot E[Y_{v,j}]\nonumber \\
E[Y_{v,i}^{t}] = p_{v}^{t} \nonumber \\
q_{v^{g}}^{t} = \sum_{i,j;i < j} p_{v}^{t} \cdot p_{v}^{t} \nonumber \\
q_{v^{g}}^{t} = \frac{b^{t} \cdot (b^{t}-1)}{2} \cdot p_{v}^{t} \cdot p_{v'}^{t} \nonumber
\end{align}

Now extending the above reasoning to any group of type $v^{g}$, a group of type $v^{g}$ can be formed if we have $n_{v,v^{g}}$ vertices of type $v$ for each $v \in v^{g}$. 

\begin{align}
q_{v^{g}}^{t} = E\Big[\prod\limits_{v\in v^{g}} \sum\limits_{i^{v}_{1},i^{v}_{2},..,i^{v}_{n_{v,v^{g}}}} Y_{v,i^{v}_{1}} \cdot Y_{v,i^{v}_{2}} \cdot .. \cdot Y_{v,i^{v}_{n_{v,v^{g}}}}\Big] \label{eq:com1} 
\end{align}

In the above expression, $i^{v}_{l} < i^{v}_{m} ~if m > l$ (Please refer to above derivation for simple case when $v^{g}=(v,v)$.)

When $i^{v}_{l} = i^{v'}_{m}$ the product of $Y_{v,i^{v}_{l}}$ and $Y_{v',i^{v'}_{m}}$ will be 0. Also $E[Y_{v,i^{v}_{l}}] = p_{v}^{t}, \forall l$, therefore, the expression in Equation \eqref{eq:com1} is equivalent to 

$$WS \cdot \prod_{v \in v^{g}} (p_{v}^{t})^{n_{v,v^{g}}}$$

where $WS$ denotes the number of ways to select the $i^{v}_{l}$ indices in Equation \eqref{eq:com1} such that the the indicator variables multiplied provide a value 1. Therefore, if group of type $v^{g} =(v_{1},v_{2},...,v_{k})$ then this is equivalent to selecting $n_{v_{1},v^{g}}$ vertices of type $v_{1}$ from $b^{t}$ followed by selecting $n_{v_{2},v^{g}}$ vertices from $b^{t}-n_{v_{1},v^{g}}$ vertices and so on. Therefore, $WS$ for $v^{g}=(v_{1},v_{2},...,v_{k})$ is computed as follows

$$WS = \frac{(b^{t})!}{(b^{t}-n_{v_{1},v^{g}})!\cdot(n_{v_{1},v^{g}})!}\cdot \frac{(b^{t}-n_{v_{1},v^{g}})!}{(b^{t}-n_{v_{1},v^{g}}-n_{v_{2},v^{g}})\cdot(n_{v_{2},v^{g}})!} \cdots\frac{(b^{t}-\sum_{i=1}^{k-1} n_{v_{i},v^{g}})!}{(n_{v_{k},v^{g}})!\cdot(b^{t}-\sum_{i=1}^{k} n_{v-{k},v^{g}})!} $$

$$WS = \frac{\prod_{i=0}^{|v^{g}|} (b^{t} - i)}{\prod_{i=1}^{k} (n_{v_{i},v^{g}})!}$$

Now for any $v^{g}$ $WS$ can be written as follows\\ 

$$WS = \frac{\prod_{i=0}^{|v^{g}|} (b^{t} - i)}{\prod_{v \in v^{g}} (n_{v,v^{g}})!}$$

Therefore, $q_{v^{g}} = \frac{\prod_{i=0}^{|v^{g}|} (b^{t} - i)}{\prod_{v \in v^{g}} (n_{v,v^{g}})!} \cdot \prod\limits_{v \in v^{g}} (p_{v}^{t})^{n_{v,v^{g}}}$

\begin{table}[t]
\center
        \begin{tabular}{|r|}
        \hline

        \begin{minipage}{0.45\textwidth}
                \vspace{0.05in}
\textbf{LPShare:}
{
\begingroup
\addtolength{\jot}{-2pt}
\begin{align}
\max & \sum_{t \in T} \sum_{u \in {\cal U}} \sum_{v^{g} \in {\cal V}^{g}} w_{u,v^{g}}^{t} \cdot x_{u,v^{g}}^{t} \\
s.t. \quad & \sum_{t'<t} \sum_{v^{g'} \in {{\cal V}^{g}}} x_{u,v^{g'}}^{t'} \cdot Pr[c_{u,v^{g'}}^{t'} > t-t'] + \nonumber\\
& \hspace{0.4in}+ \sum_{v^{g} \in {\cal V}^{g}} x_{u,v^{g}}^{t} \leq  1 ::: \forall u,t \label{cons:opt-c1}\\
& \sum_{v^{g}; v \in v^{g}} \sum_{u \in {\cal U}} n_{v,v^{g}} \cdot x_{u,v^{g}}^{t} \leq q_{v}^{t} ::: \forall v,t \label{cons:opt-c2}\\
& \sum_{u \in {\cal U}} x_{u,v^{g}}^{t} \leq q_{v^{g}}^{t} ::: \forall v^{g},t \label{cons:opt-c3}\\
& 0 \leq x_{u,v^{g}}^{t} \leq 1 ::: \forall u,v^{g},t \label{cons:opt-c4}
\end{align}
\endgroup }
\end{minipage} \\
        \hline
        \end{tabular}

        \caption{Optimization Formulation - Multi-Capacity}
        \label{cap2:opt1}
        \end{table}
\section{Details about the Optimization Formulation in Table \ref{cap2:opt1}}
Let the number of vertices of type $v$ available in round $t$ is $m_{v}^{t}$. We use $m_{v^{g}}^{t}$ to denote the number of groups of type $v^{g}$ which can be formed in round $t$. $n_{v,v^{g}}$ denotes the number of vertices of type $v$ present in the group of type $v^{g}$. 
Let $x^{t}_{u,v^{g}}$ denotes the assignment of resource $u$ to the group of type $v^{g}$. Let $y^{t}_{v,v^{g}}$ denotes the flow on the edge from $v$ to $v^{g}$ where $v$ is a part of the group of type $v^{g}$. Therefore, we will have following flow preservation constraints:

\begin{align}
&\sum_{t'}\sum_{v^{g'} \in {\cal V}^{g}} x^{t}_{u,v^{g'}} \cdot Pr(c_{u,v^{g'}}^{t'} > t-t') +\sum_{v^{g} \in {\cal V}^{g}} x_{u,v^{g}}^{t} \leq 1 ::: \forall u,t \label{eq:11} \\
&\sum_{u \in {\cal U}} x_{u,v^{g}}^{t} \leq m_{v^{g}}^{t} ::: \forall v^{g},t \label{eq:12}\\
&\sum_{v^{g}; v \in v^{g}} y_{v,v^{g}}^{t} \cdot n_{v,v^{g}} \leq m_{v}^{t} ::: \forall v,t \label{eq:13}\\
&\sum_{v \in v^{g}} n_{v,v^{g}} \cdot y_{v,v^{g}}^{t} \leq sizeof(v^{g}) \cdot m_{v^{g}}^{t} ::: \forall v^{g},t \label{eq:14}\\
&y_{v,v^{g}}^{t} = \sum_{u \in {\cal U}} x_{u,v^{g}}^{t} ::: \forall v^{g}; v \in v^{g}, t  \label{eq:15}
\end{align}
The Equation \eqref{eq:15} is the conditional equal flow constraint which states that the total incoming flow to a group will be equal to the total outgoing flow to each of the vertex which is a part of the group. 

Substituting Equation \eqref{eq:15} in Equation \eqref{eq:13}, we get 
\begin{align}
& \sum_{v^{g}; v \in v^{g}} \sum_{u \in {\cal U}} x_{u,v^{g}}^{t} \leq m_{v}^{t} ::: \label{eq:16} \forall v,t
\end{align}

As $m_{v^{g}}^{t} \geq min(m_{v}^{t})$, if $m_{v}^{t}$ is an integer, therefore, Equation \eqref{eq:12} is redundant in the presence of Equation \eqref{eq:16}.

Similarly on substituting Equation \eqref{eq:15} in Equation \eqref{eq:14}, we get, 
\begin{align}
& \sum_{v \in v^{g}} n_{v,v^{g}} \sum_{u \in {\cal U}} x_{u,v^{g}}^{t} \leq sizeof(v^{g}) \cdot m_{v^{g}}^{t} :::  \forall v^{g},t\\
& \sum_{u \in {\cal U}} x_{u,v^{g}}^{t} \leq m_{v^{g}}^{t} ::: \label{eq:17} \forall v^{g},t
\end{align}

This is same as Equation \eqref{eq:12} which is redundant. 

Therefore, we get the optimization formulation provided in Table \ref{cap2:opt1_2}. Please note that we keep the redundant constraint in the optimization formulation.

We replace $m_{v^{g}}^{t}$ and $m_{v}^{t}$ by $q_{v^{g}}^{t}$ and $q_{v}^{t}$ which denote the expected number of groups/vertices. The equations remains same as Equations \eqref{eq:11} - \eqref{eq:15}. 

But unlike earlier case, we can not say that Equation \eqref{eq:12} is redundant in presence of Equation \eqref{eq:16}.

This is because $q_{v}^{t}$ can lie between 0 and 1. 

Therefore, we get the optimization formulation presented in Table \ref{cap2:opt1}.

\section{Proof of Proposition 3}
\begin{prop}
The optimal value of the LP presented in Table \ref{cap2:opt1} is a valid upper bound on the offline optimal value. 
\end{prop}
\textbf{Proof:}
The proof is similar to the proof of Proposition 1. 

To show that the LP provides a valid upper bound on the offline optimal solution, we prove that the expected value of matching is less than or equal to the optimal solution of LP. We have two distributions, one corresponding to the arrival of vertices ($p_{v}^{t}$) and another corresponding to the number of rounds of unavailability ($c_{u,v^{g}}^{t}$). To prove that the optimal value of the LP provides a valid upper bound in presence of both distributions, similar to earlier works \cite{karthik2019}, we need to make an assumption that the optimal assignment only depends on the arrival distribution and is independent of the value of $c_{u,v^{g}}^{t}$~\footnote{Instead of this, we can also assume that $c_{u,v^{g}}^{t}$ is a known constant which is a weaker assumption than assuming that the optimal assignment is independent of distribution of the number of rounds of unavailability. The proof is similar for both assumptions.}.

Consider a realization of arrivals denoted by sequence $a$. Let $m_{v}^{t}(a)$ denotes the number of vertices of type $v$ arriving in round $t$ for arrival sequence $a$ and $m_{v^{g}}^{t}(a)$ denotes the number of groups of type $v^{g}$ which can be formed in round $t$ in the arrival sequence $a$. Similarly, consider a realization of the number of rounds of unavailability denoted by sequence $r$, where $c_{u,v^{g}}^{t}(r)$ denotes the number of rounds for which resource $u$ becomes unavailable on matching with group of type $v^{g}$ in sequence $r$. $\delta(c_{u,v^{g}}^{t'}(r) > t-t')$ is an indicator variable denoting that the number of rounds for which resource $u$ becomes unavailable on being assigned to group of type $v^{g}$ in round $t'$ is greater than $t-t'$. Now, for any arrival sequence $a$ and any realization of the number of rounds of unavailability $r$, the offline solution can be computed by solving the optimization program in Table \ref{cap2:opt1_2}~\footnote{We index each of $m_{v}^{t}$, $m_{v^{g}}^{t}$, $x_{u,v}^{t}$  by $a$ and $c_{u,v^{g}}^{t}$ by $r$ to denote the instance for sequence $a$ and $r$.}.

As mentioned before, we make an assumption that the optimal assignment is independent of the realization of the number of rounds of unavailability, i.e., it only depends on $a$ and not $r$. Therefore, we use $x^{*}(a)$ to denote the optimal solution of the formulation in Table \ref{cap2:opt1_2} for sequence $a$ and $r$. The expected number of times $(u,v^{g})$ is matched at $t$ is given by $\sum_{a} x^{*,t}_{u,v^{g}}(a) \cdot P(a)$. To prove that, the optimal value of LP in Table \ref{cap2:opt1} is a valid upper bound on the optimal solution, We show that $\forall u,v^{g},t ~~\sum_{a}x^{*,t}_{u,v^{g}}(a) \cdot P(a)$ is a feasible solution to the LP.

The optimization in Table \ref{cap2:opt1_2} is solved for each sequence $a$ and $r$. As we used $x^{*}(a)$ to denote the optimal solution for the sequence $a$ and $r$, therefore, $x^{*}(a)$ satisfies the constraints of formulation in Table \ref{cap2:opt1_2}. Hence, we get

\begin{align}
& \sum_{t'<t} \sum_{v^{g'} \in {\cal V}^{g}} x^{*,t'}_{u,v^{g'}}(a) \cdot \delta(c_{u,v^{g'}}^{t'}(r) > t-t') + \sum_{v^{g} \in {\cal V}^{g}} x^{*,t}_{u,v^{g}}(a) \leq  1 ::: \forall u,t \label{cons:optc13_a}\\
&\sum_{v^{g}; v \in v^{g}} \sum_{u \in {\cal U}} n_{v,v^{g}} \cdot x^{*,t}_{u,v^{g}}(a) \leq m_{v}^{t}(a) ::: \forall v,t \label{cons:optc14_a}\\
&\sum_{u \in {\cal U}} x^{*,t}_{u,v^{g}}(a) \leq m_{v^{g}}^{t}(a) ::: \forall v^{g},t \label{cons:optc15_a}
\end{align}

Multiplying Equations \eqref{cons:optc13_a},\eqref{cons:optc14_a} and \eqref{cons:optc15_a} by $P(a) \cdot P(r)$, probability of sequence $a$ and $r$, and performing summation, we get

\begin{align}
& \sum_{a} \sum_{r}\sum_{t'<t}  \sum_{v^{g'} \in {\cal V}^{g}} x^{*,t'}_{u,v^{g'}}(a) \cdot \delta(c_{u,v^{g'}}^{t'}(r) > t-t') \cdot P(a) \cdot P(r) + \nonumber \\ 
&\hspace{1in} \sum_{a}\sum_{r} \sum_{v^{g} \in {\cal V}^{g}} x^{*,t}_{u,v^{g}}(a) \cdot P(a) \cdot P(r) \leq  \sum_{a} \sum_{r} 1 \cdot P(a) \cdot P(r)  ::: \forall u,t \label{cons:optc13_a1}\\
&\sum_{a} \sum_{r}\sum_{v^{g}; v \in v^{g}} \sum_{u \in {\cal U}} n_{v,v^{g}} \cdot x^{*,t}_{u,v^{g}}(a) \cdot P(a) \cdot P(r) \leq \sum_{a} \sum_{r} m_{v}^{t}(a) \cdot P(a) \cdot P(r) ::: \forall v,t \label{cons:optc14_a1}\\
&\sum_{a} \sum_{r}\sum_{u \in {\cal U}} x^{*,t}_{u,v^{g}}(a) \cdot P(a) \cdot P(r) \leq \sum_{a} \sum_{r} m_{v^{g}}^{t}(a) \cdot P(a) \cdot P(r) ::: \forall v^{g} \label{cons:optc15_a1}
\end{align}

$\sum_{a} \sum_{r} 1 \cdot P(a) \cdot P(r) = 1$ and $\sum_{a} \sum_{r} m_{v}^{t}(a) \cdot P(a) \cdot P(r) = \sum_{a} m_{v}^{t}(a) \cdot P(a)$ denotes the expected number of vertices of type $v$ arriving in round $t$ and $\sum_{a} \sum_{r} m_{v^{g}}^{t}(a) \cdot P(a) \cdot P(r) = \sum_{a} m_{v^{g}}^{t}(a) \cdot P(a)$ denotes the expected number of groups of type $v^{g}$ formed in round t. Therefore, 

$$\sum_{a} m_{v}^{t}(a) \cdot P(a) = q_{v}^{t}$$. 
$$\sum_{a} m_{v^{g}}^{t}(a) \cdot P(a) = q_{v^{g}}^{t}$$. 

Also, as $x^{*}$ is independent of $r$, therefore,

$$\sum_{a} \sum_{r} x^{*,t}_{u,v^{g}}(a) \cdot  P(a) \cdot P(r) = \sum_{a} x^{*,t}_{u,v^{g}}(a) \cdot P(a)$$

On substituting these values, we get,
\begin{align}
& \sum_{a} \sum_{t'<t} \sum_{v^{g'} \in {\cal V}^{g}} x^{*,t}_{u,v^{g'}}(a) \cdot P(a) \sum_{r} \delta(c_{u,v^{g'}}^{t'}(r) > t-t')\cdot P(r) + \sum_{a} \sum_{v^{g} \in {\cal V}^{g}} x^{*,t}_{u,v^{g}}(a) \cdot P(a) \leq  1  ::: \forall u,t \label{cons:optc13_a11}\\
&\sum_{a} \sum_{v^{g}; v \in v^{g}} \sum_{u \in {\cal U}} n_{v,v^{g}} \cdot x^{*,t}_{u,v^{g}}(a) \cdot P(a) \leq q_{v}^{t} ::: \forall v,t \label{cons:optc14_a12}\\
&\sum_{a} \sum_{u \in {\cal U}} x^{*,t}_{u,v^{g}}(a) \cdot P(a) \leq q_{v^{g}}^{t} ::: \forall v^{g},t \label{cons:optc15_a13}
\end{align}

As,

$$\sum_{r} \delta(c_{u,v^{g'}}^{t'}(r) > t-t') \cdot P(r) = Pr(c_{u,v^{g'}}^{t'} > t-t') $$

On substituting these values, Equations \eqref{cons:optc13_a11},\eqref{cons:optc14_a12} and \eqref{cons:optc15_a13} become similar to the constraints of the optimization formulation in Table \ref{cap2:opt1} with $x_{u,v^{g}}^{t} = \sum_{a} x^{*,t}_{u,v^{g}}(a) \cdot P(a)$. Therefore, $\sum_{a} x^{*}(a) \cdot P(a)$ is a feasible solution to LP in Table \ref{cap2:opt1}. Hence it is proved that the LP provides a valid upper bound on the optimal solution.

\begin{table}[t]
\center
        \begin{tabular}{|r|}
        \hline

        \begin{minipage}{0.95\textwidth}
                \vspace{0.05in}
{
\begingroup
\addtolength{\jot}{-2pt}
\begin{align}
\max & \sum_{t \in T} \sum_{u \in {\cal U}} \sum_{v^{g} \in {\cal V}^{g}} w_{u,v^{g}}^{t} \cdot x_{u,v^{g}}^{t}(a) \\
s.t. \quad & \sum_{t'<t} \sum_{v^{g'} \in {\cal V}^{g}} x_{u,v^{g'}}^{t'}(a) \cdot \delta(c_{u,v^{g'}}^{t'}(r) > t-t') + \sum_{v^{g} \in {\cal V}^{g}} x_{u,v^{g}}^{t}(a) \leq  1 ::: \forall u,t \label{cons:optc13}\\
&\sum_{v^{g}; v \in v^{g}} \sum_{u \in {\cal U}} n_{v,v^{g}} \cdot x_{u,v^{g}}^{t}(a) \leq m_{v}^{t}(a) ::: \forall v,t \label{cons:optc14}\\
&\sum_{u \in {\cal U}} x_{u,v^{g}}^{t}(a) \leq m_{v^{g}}^{t}(a) ::: \forall v^{g} \label{cons:optc15}\\
& x_{u,v^{g}}^{t}(a) \in \{0,1\} ::: \forall u, v^{g}, t
\end{align}
\endgroup }
\end{minipage} \\
        \hline
        \end{tabular}
        \caption{ Optimization Formulation - Multi Capacity - For a fixed sequence $a$ and $r$}
        \label{cap2:opt1_2}
        \end{table}

\section {Example showing the groups considered at different steps}
\begin{Example}
\label{ex:1}
Suppose ${\cal V}=\{v_1,v_2 \}$, $b^t =3$. Out of the three incoming vertices - two vertices are of type $v_1$ and one vertex is of type $v_2$. To distinguish between two vertices of type $v_1$, we refer them by $v_1(1)$ and $v_1(2)$. On random shuffling of these three vertices, they are present in the following order:
\begin{center}
Sequence : ($v_1(1) , v_2, v_1(2)$)
\end{center}
Step $(1,2)$ represent that the first and second vertex in the above sequence is considered. We define an ordering over types of vertices. In this example, let $v_1$ > $v_2$. So, whenever we are considering group formed with these two types of vertices, we will always consider $(v_1, v_2)$ and not $(v_2, v_1)$. Therefore, in this example, we will consider the group at $(1,2)$ and $(3,2)$ not at $(2,1)$ or $(2,3)$. This ensures that we are processing each group only once.

\begin{table}[h]
\center
       \begin{tabular}{|c|c|c|c|c|c|}
\hline
Step & $v^{g}$ & Step & $v^{g}$ &Step & $v^{g}$\\
\hline
(1,1) & $v_{1}(1)$ & (2,1) & \st{$(v_{2},v_{1}(1))$} & (3,1) & \st{$(v_{1}(2),v_{1}(1))$}\\
\hline
(1,2) & $(v_{1}(1),v_2)$ & (2,2) & $v_{2}$ & (3,2) & $(v_{1}(2),v_{2})$\\
\hline
(1,3) & $(v_{1}(1),v_{1}(2))$ & (2,3) & \st{$(v_{2},v_{1}(2))$} & (3,3) & $v_{1}(2)$\\
\hline
\end{tabular}
\vspace{-0.15in}
\end{table}

In another arrival sequence, two vertices are of type $v_2$ and one vertex is of type $v_1$. To distinguish between these two vertices of type $v_2$, we refer to them by $v_2(1)$ and $v_2(2)$. On random shuffling of these three vertices, they are present in following order:

\begin{center}
Sequence : ($v_2(1) , v_1, v_2(2)$)
\end{center}

In this case the groups will be processed as shown in the below table,
\begin{table}[h]
\center
        \begin{tabular}{|c|c|c|c|c|c|}
\hline
Step & $v^{g}$ & Step & $v^{g}$ &Step & $v^{g}$\\
\hline
(1,1) & $v_{2}(1)$ & (2,1) & $(v_{1},v_{2}(1))$ & (3,1) & \st{$(v_{2}(2),v_{2}(1))$}\\
\hline
(1,2) & \st{$(v_{2}(1),v_{1})$} & (2,2) & $v_{1}$ & (3,2) & \st{$(v_{2}(2),v_{1})$}\\
\hline
(1,3) & $v_{2}(1),v_{2}(2))$ & (2,3) & $(v_{1},v_{2}(2))$ & (3,3) & $v_{2}(2)$\\
\hline
\end{tabular}
\vspace{-0.15in}
\end{table}

So across these 2 sequences, we can see that the group of type $(v_1, v_2)$ is processed at 4 places: $(1,2) (2,1),(2,3) (3,2)$ – Similarly if we create more sequences, we will observe that the group of type $(v_1, v_2)$ can be considered at 6 places (all places except $(1,1) (2,2) (3,3)$).

\end{Example}

\section{Complete proof of Proposition 4}
\label{sect:complete}

\noindent {\bf Proving $\beta_{u,(i,j)}^{t} \geq 1-\gamma$:} 

To prove that $\beta_{u,(i,j)}^{t} \geq 1-\gamma$, we use mathematical induction, at $t=1$, initially all $u$ are available, therefore, 

$$\beta_{u,(1,1)}^{1}=1$$

Please note that $\beta_{u,(i,j)}^{t}$ keeps on decreasing as $i,j$ increases for fixed $u$ and $t$, therefore, we only show for the value of $\beta_{u,(b^{1},b^{1})}^{1}$.
\begin{align}
\beta_{u,(b^{1},b^{1})}^{1} & = 1 - \sum_{v^{g}}\sum_{i=1}^{b^{1}-1}\sum_{j=1}^{b^{1}} \frac{x^{*,1}_{u,v^{g}} \cdot \gamma}{h_{v^{g}}^{1} \cdot P_{v^{g},(i,j)}^{1} \cdot \beta_{u,(i,j)}^{1}} \cdot P_{v^{g},(i,j)}^{1} \cdot \beta_{u,(i,j)}^{1} - \sum_{v^{g}} \sum_{j=1}^{b^{1}-1} \frac{x^{*,1}_{u,v^{g}} \cdot \gamma}{h_{v^{g}}^{1} \cdot P_{v^{g},(i,j)}^{1} \cdot \beta_{u,(i,j)}^{1}} \cdot P_{v^{g},(i,j)}^{1} \cdot \beta_{u,(i,j)}^{1}
\end{align}

As mentioned before in ADAPShare-$\kappa$ description, each group will be considered for assignment at $h_{v^{g}}^{t}$ steps, and at step $(b^{t},b^{t})$, a single vertex will be considered, therefore, 

\begin{align}
\beta_{u,(b^{1},b^{1})}^{1} & \geq 1 - \sum_{v^{g}} (x^{*,1}_{u,v^{g}} \cdot \gamma)
\end{align}

As maximum value of $\sum_{v^{g}} x^{*,t}_{u,v^{g}}$ is 1. Therefore, $\beta_{u,(b^{1},b^{1})}^{1} \geq 1-\gamma$. 

As it is valid for $t=1$, we prove it by induction. Assume that it is valid for all $t' < t$, i.e., in online case the resource $u$ is matched $x^{*,t}_{u,v^{g}} \cdot \gamma$ times in round $t$ to group of type $v^{g}$. Therefore,

\begin{align} 
\beta_{u,(1,1)}^{t} & = 1 - \sum_{v^{g'}} \sum_{t' < t} x^{*,t'}_{u,v^{g'}} \cdot \gamma \cdot Pr(c_{u,v^{g}}^{t'} \geq t-t') 
\end{align} 

From Equation \eqref{cons:opt-c1} of the optimization program in Table \ref{cap2:opt1}, we have 

$$\sum_{v^{g'}} \sum_{t' < t} (x^{*,t'}_{u,v^{g'}} \cdot \gamma \cdot Pr(c_{u,v^{g'}}^{t'} > t - t')) \leq \gamma - \sum_{v^{g}} x^{*,t}_{u,v^{g}} \cdot \gamma$$

Multiplying both sides of the above equation by -1, we get

$$-1 \cdot \sum_{v^{g'}} \sum_{t' < t} (x^{*,t'}_{u,v^{g'}} \cdot \gamma \cdot Pr(c_{u,v^{g'}}^{t'} > t - t')) \geq -1 \cdot (\gamma - \sum_{v^{g}} x^{*,t}_{u,v^{g}} \cdot \gamma)$$

Adding 1 on both sides in above equation, we get

$$1 -\sum_{v^{g'}} \sum_{t' < t} (x^{*,t'}_{u,v^{g'}} \cdot \gamma \cdot Pr(c_{u,v^{g'}}^{t'} > t - t')) \geq 
1 -\gamma + \sum_{v^{g}} x^{*,t}_{u,v^{g}} \cdot \gamma$$

Therefore, 

$$\beta_{u,(1,1)}^{t} \geq  1 -\gamma + \sum_{v^{g}} x^{*,t}_{u,v^{g}} \cdot \gamma \geq 1 - \gamma$$
\begin{align}
\beta_{u,(b^{t},b^{t})}^{t} & = \beta_{u,(1,1)}^{t} - \sum_{v^{g}}\sum_{i=1}^{b^{t}-1}\sum_{j=1}^{b^{t}} \frac{x^{*,t}_{u,v^{g}} \cdot \gamma}{h_{v^{g}}^{t} \cdot P_{v^{g},(i,j)}^{t} \cdot \beta_{u,(i,j)}^{t}} \cdot P_{v^{g},(i,j)}^{t} \cdot \beta_{u,(i,j)}^{t} \nonumber \\ 
\hspace{1in} & - \sum_{v^{g}} \sum_{j=1}^{b^{t}-1} \frac{x^{*,t}_{u,v^{g}} \cdot \gamma}{h_{v^{g}}^{t} \cdot P_{v^{g},(i,j)}^{t} \cdot \beta_{u,(i,j)}^{t}} \cdot P_{v^{g},(i,j)}^{t} \cdot \beta_{u,(i,j)}^{t} 
\end{align}
Similar to $t=1$ case 
\begin{align}
\beta_{u,(b^{t},b^{t})}^{t} & = \beta_{u,(1,1)}^{t} - \sum_{v^{g}} (x^{*,t}_{u,v^{g}} \cdot \gamma) \nonumber \\ 	
\beta_{u,(b^{t},b^{t})}^{t} & \geq 1 -\gamma + \sum_{v^{g}} x^{*,t}_{u,v^{g}} \cdot \gamma - \sum_{v^{g}} (x^{*,t}_{u,v^{g}} \cdot \gamma) \\
\beta_{u,(b^{t},b^{t})}^{t} &\geq 1- \gamma
\end{align}

Therefore, $$\beta_{u,(i,j)}^{t} \geq (1-\gamma) , \forall t,i,j$$

\section{Proof of Corollary 1}
\begin{prop}
The online algorithm ADAPShare is $\gamma$ competitive (The value of $\gamma$ is the solution to the equation $(1-\gamma)^{(\kappa+1)} = \gamma$).
\end{prop}

\textbf{Proof:} Using Proposition 5(in the paper), we can show that ADAPShare-$\kappa$ is $\gamma$ competitive. Therefore, we need to find the maximum value of $\gamma$ for which the assignment rule for ADAPShare-$\kappa$ is valid. We highlight the differences in comparison to proof of Proposition 4. 

We can now group $\kappa$ vertices together from the  $b^{t}$ vertices arriving in round $t$, therefore, we define $(b^{t})^{\kappa}$ steps in our algorithm. 

Similar to ADAPShare-2, Step $(i_{1},i_{2},...,i_{\kappa})$ ($i_{1} \neq i_{2} \neq ...\neq i_{\kappa}$) denotes that the group formed by vertex at $i_{1}^{th},i_{2}^{th}...$ and $..i_{\kappa}^{th}$ label is considered. Step $(i,i,..,i)$ denotes that a group of size 1 with one vertex at $i^{th}$ position is considered. Similarly, we can define steps where groups of size $2$ to $\kappa-1$ are considered. Group of size $s$ will be considered at $\prod\limits_{i=0}^{s-1}(b^{t} -i)$ steps. There will be $(b^{t})^{\kappa} - \sum\limits_{s=1}^{\kappa} \prod\limits_{i=0}^{s} (b^{t}-i)$ steps where algorithm does not do anything, i.e., none of the groups is considered for assignment at these steps. 

Similar to ADAPShare-2 case, we find the maximum value of  
$\frac{\prod\limits_{v\in v^{g}} (p_{v}^{t})^{n_{v,v^{g}}}}{P_{v^{g},(i_{1},i_{2},..,i_{\kappa})}^{t}} $ and then show that $\beta_{u,(i_{1},i_{2},...,i_{\kappa})}^{t} \geq \frac{\gamma \cdot \prod\limits_{v\in v^{g}} (p_{v}^{t})^{n_{v,v^{g}}}}{P_{v^{g},(i_{1},i_{2},..,i_{\kappa})}^{t}}$ for this maximum value.

Please note that 

$P_{v^{g},(i_{1},i_{2},..,i_{\kappa})}^{t} = \prod\limits_{ v \in v^{g}} P_{v,i_{v},(i_{1},i_{2},..,i_{\kappa})}^{t} $. Therefore, we compute $P_{v,i_{v},(i_{1},i_{2},..,i_{\kappa}) }^{t} \forall v$  where $P_{v,i_{v},(i_{1},..,i_{\kappa})}^{t}$ denotes the probability that vertex of type $v$ labeled as $i_{v}^{th}$ vertex is available at step $(i_{1},..,i_{\kappa})$. Please note that $P_{v,i,(1,1,..,1)}^{t} = p_{v}^{t} \forall i $.\\

{ 
\begin{align}
& P_{v,i_{v},(i_{1},..,i_{\kappa})}^{t}  = p_{v}^{t} - \nonumber\\ 
& \sum_{u} \sum_{v^{g'}; v \in v^{g'}}  \sum_{j_{1}=1}^{i^{1}-1}..\sum_{j_{\kappa}=1}^{i_{\kappa}-1} \frac{x^{t}_{u,v^{g'}} \cdot \gamma}{h_{v^{g}}^{t} \cdot P_{v^{g'},(j_{1},..,j_{\kappa})}^{t} \cdot \beta_{u,(j_{1},..,j_{\kappa})}^{t}} \cdot P_{v^{g'},(j_{1},..,j_{\kappa})}^{t} \cdot \beta_{u,(j_{1},..,j_{\kappa})}^{t} \label{eq:mvg} 
\end{align}
}

Now, similar to $\kappa=2$ case, in the above equation we only consider the steps where vertex of type $v$ has label $i_{v}$. 
Let $e_{v^{g},v,i_{v},(i_{1},..,i_{\kappa})}^{t}$ denote the maximum number of steps (for group of type $v^{g}$) which can affect the computation of $P_{v,i_{v},(i_{1},..,i_{\kappa})}^{t}$, then
\begin{align}
& P_{v,i_{v},(i_{1},..,i_{\kappa})}^{t}  = p_{v}^{t} - \nonumber\\ 
& \sum_{u} \sum_{v^{g'};v \in v^{g'}}  e_{v^{g},v,i_{v},(..)}^{t} \cdot \frac{x^{t}_{u,v^{g'}} \cdot \gamma}{h_{v^{g}}^{t}}  
\end{align}
And 
$e_{v^{g},v,i_{v},(i_{1},..,i_{\kappa})}^{t}=\frac{\prod_{i=1}^{|v^{g}|} (b^{t}-i)}{(\prod_{v' \in v^{g}; v\neq v'} (n_{v',v^{g}})!) (n_{v,v^{g}} -1)!}$ 

Therefore, 
$\frac{e_{v^{g},v,i_{v},(i_{1},..,i_{\kappa})}^{t}}{h_{v^{g}}^{t}} =  \frac{n_{v,v^{g}}}{b^{t}}$

Therefore, we can get following by proceeding in similar way as the analysis for $\kappa=2$



\begin{align}
P_{v,i_{v},(i_{1},..,i_{\kappa})}^{t} & \geq p_{v}^{t} - \gamma  \cdot  p_{v}^{t}\\
P_{v,i_{v},(i_{1},..,i_{\kappa})}^{t} & \geq (1-\gamma) \cdot p_{v}^{t}
\end{align}


Similar to $\kappa=2$, we can use mathematical induction to show $\beta_{u,(i_{1},..,i_{\kappa})}^{t} \geq \frac{\gamma}{(1-\gamma)^{\kappa}}$ 
Therefore, maximum value of $\gamma$ for which assignment rule is valid is the solution to the equation, $\gamma = (1-\gamma)^{\kappa+1}$


\begin{algorithm}
\caption{ADAPShare-$\kappa$($\gamma$)}
\begin{algorithmic}[1]
\FOR{$t < T$}
\STATE{Generate $b^{t}$ uniform random numbers and sort the vertices in order of generated random numbers. Label the vertices from 1 to $b^{t}$.}
\STATE{$i_{1}=1,i_{2}=1,...,i_{\kappa} =1$}
\WHILE{$i_{1} \leq b^{t} || i_{2} \leq b^{t} ||...|| i_{\kappa} \leq b^{t} $}
\STATE{$v^{g}$ = group formed at step $i_{1},i_{2},...,i_{\kappa}$ based on the labels assigned to the vertices.}
\IF{$v^{g}$ is a valid group}
\STATE{If ${\cal E}_{*,v^{g},t} != \phi$, then choose $(u,v^{g}) \in {\cal E}_{*,v^{g},t}$ with probability $p$ where $p= \frac{x^{*,t}_{u,v^{g}} \cdot \gamma}{h_{v^{g}}^{t} \cdot P_{v^{g},(i_{1},i_{2},..,i_{\kappa})}^{t} \cdot \beta_{u,(i_{1},..,i_{\kappa})}^{t}}$}
\STATE{Update ${\cal E}_{*,*,t}$,available groups based on the group considered in previous steps. }

\ENDIF
\STATE{Increment the step $i_{1}=1,i_{2}=1,...,i_{\kappa}$}
\ENDWHILE
\ENDFOR
\end{algorithmic}
\label{alg:suggestmatchhighcap}
\end{algorithm}

\end{document}